\documentclass[journal]{IEEEtran}
\newcommand{\etal}{\textit{et al.}}
\usepackage{graphicx}
\usepackage{enumitem}
\usepackage{booktabs}
\usepackage{array}
\usepackage{xcolor}
\usepackage{multirow}
\usepackage{multicol}
\usepackage{subfigure}
\usepackage{amsmath}

\begin{document}

\title{Action Quality Assessment using Siamese Network-Based Deep Metric Learning}

\author{Hiteshi Jain, Gaurav Harit and Avinash Sharma
\thanks{Hiteshi Jain and Gaurav Harit are with Department of Computer Science and Engineering, Indian Institute of Technology Jodhpur, Karwar, Jodhpur, India, 342037
 e-mail: jain.4@iitj.ac.in.}
\thanks{Avinash Sharma is from International Institute of Information Technology, Gachibowli, Hyderbad,  500032, India} }

\markboth{}%
{Shell \MakeLowercase{\textit{et al.}}: Bare Demo of IEEEtran.cls for IEEE Journals}

\maketitle

\begin{abstract}
Automated vision-based score estimation models can be used as an alternate opinion to avoid judgment bias. 
In the past works the score estimation models were learned by regressing the video representations to the ground truth score provided by the judges.
However such regression-based solutions lack interpretability in terms of giving reasons for the awarded score. 
One solution to make the scores more explicable is to compare the given action video with a reference video.
This would capture the temporal variations vis-\'a-vis the reference video and map those variations to the final score.
In this work, we propose a new action scoring system as a two-phase system: (1) A Deep Metric Learning Module that learns similarity between any two action videos based on their ground truth scores given by the judges; (2) A Score Estimation Module that uses the first module to find the resemblance of a video to a reference video in order to give the assessment score.
%
%
The proposed scoring model has been tested for Olympics Diving and Gymnastic vaults and the model outperforms the existing state-of-the-art scoring models.
\end{abstract}

\begin{IEEEkeywords}
automatic scoring, Siamese, LSTM, Deep Metric Learning
\end{IEEEkeywords}

\IEEEpeerreviewmaketitle

\begin{figure*}[t!]
\centering
\subfigure[][]{%
\includegraphics[scale=0.5]{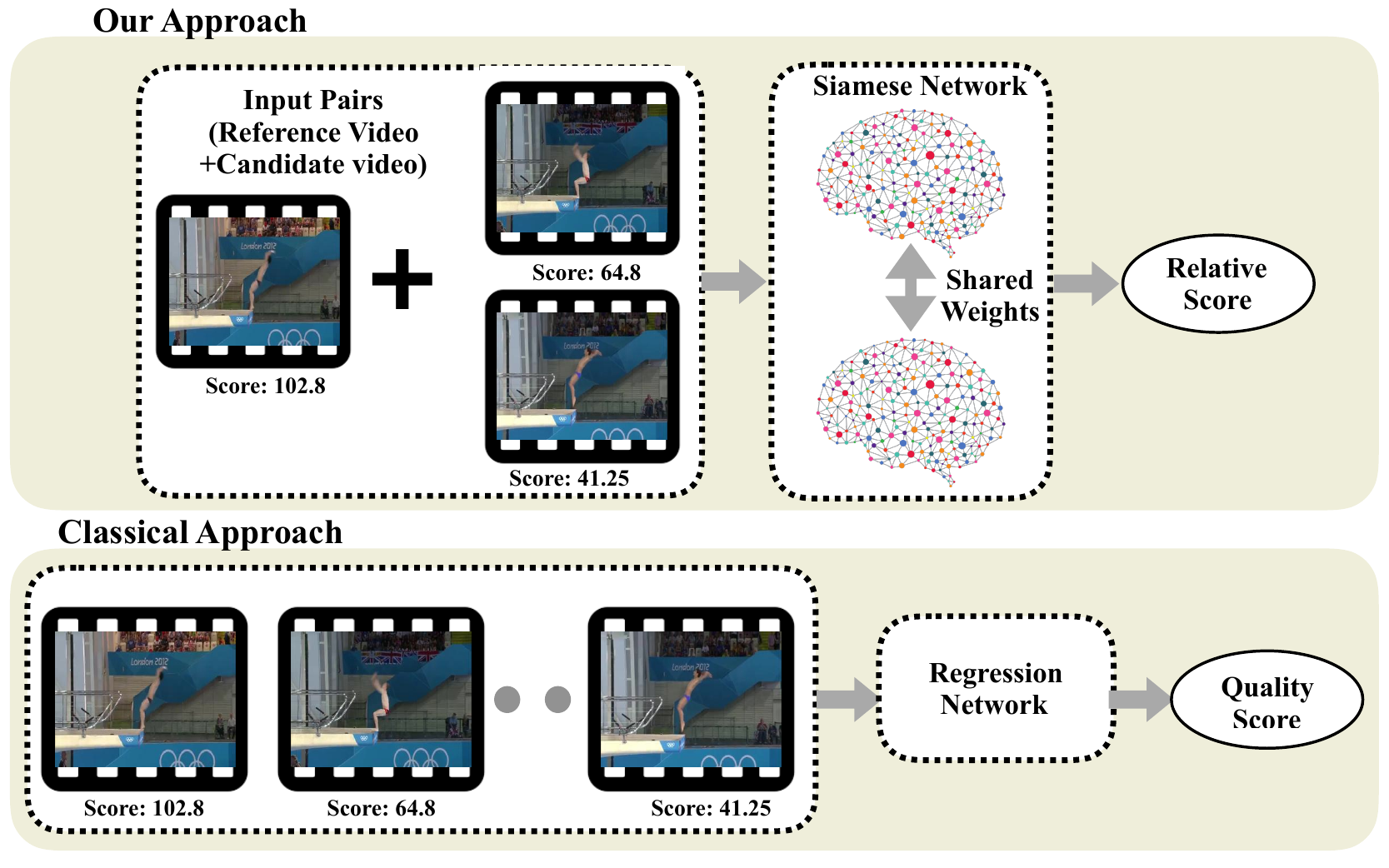}}%
\hspace{4pt}%
\subfigure[][]{%
\includegraphics[scale=0.5]{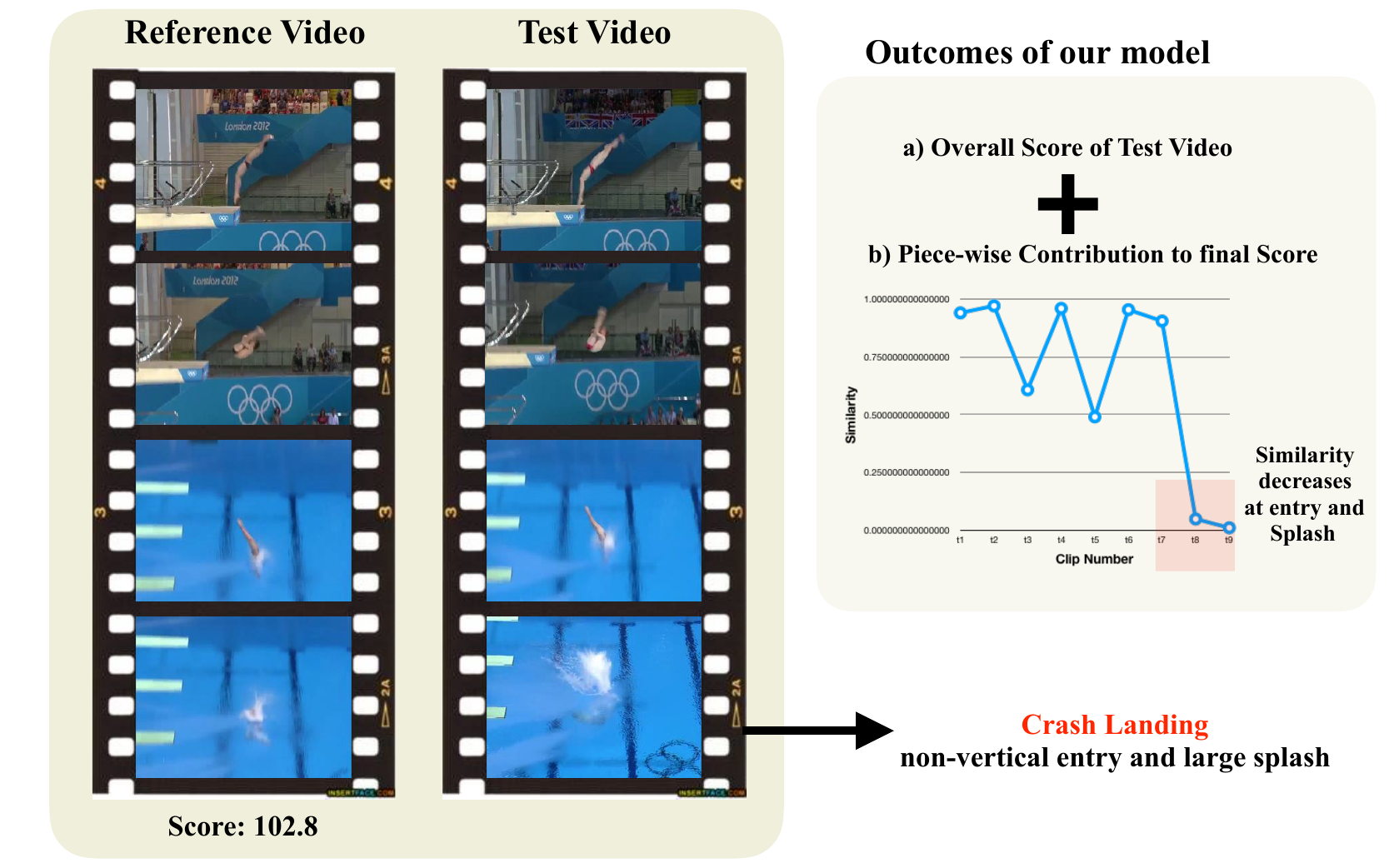}} \caption{Overview of our work; The task of scoring is transformed to the task of comparing a performance to an expert performance; Outputs of our model: Overall performance score and Clip-level feedback}
\label{fig:cover}%
\end{figure*}

\section{Introduction}

\IEEEPARstart{H}{umans} strive to attain perfection and efficiency in their day-to-day activities and maximize the outcome of their actions. They improve their performances by comparing themselves with others or seeking feedback from experts. The task of Action Quality Assessment (AQA) involves quantification of the quality of action \textit{i.e.} determining \textit{how well} an action was performed.  Unlike the traditional action recognition/classification task widely attempted in the computer vision community, where a subset of discriminative frames can help identify the type of action, the task of action assessment is more challenging as it requires quantitative and qualitative assessment of the entire action sequence. AQA has applications in many areas such as sports \cite{pirsiavash2014assessing,parmar2017learning}, health-care \cite{zia2015automated}, rehabilitation \cite{parmar2016measuring}, exercise \cite{jain2016framework,jain2018detecting} and actions of daily living \cite{doughty2018s}.


The key challenge in AQA is modelling the human subjectivity as this task suffers from significant influence of human (or expert) bias. Traditionally, the subjectivity in action assessment has been addressed by considering feedbacks given by multiple experts, \textit{e.g.} Olympics events allows averaging scores from multiple judges (while dropping the highest and lowest scores).  Such biases may include nationalistic bias \cite{ansorge1988international,emerson2009assessing}, where judges give higher scores to performers from their own countries, a difficulty bias \cite{morgan2014harder}, where athletes attempting more difficult routines receive higher execution scores, etc. Nevertheless, the human bias in scoring is inevitable and hence only partially addressable in any scenario. Thus, an automated AQA system should aim bringing more interpretability in predicted scores while being an objective source of alternate evaluation. However, such a solution needs to learn all possible nuances that a human expert/judge learns over the years. 


 Majority of the existing works on action quality assessment/scoring ({\it e.g.,}\cite{pirsiavash2014assessing,venkataraman2015dynamical,parmar2017learning,xiang2018s3d}) perform holistic evaluation of action videos and learn to predict the assessment score from the video features. This is achieved by modeling the task of score estimation as a  regression task where the action video representation is regressed to the ground truth score provided by the judges. Apart from holistic evaluation, it is also important to have a feedback on how different sub-actions contribute to the final score for better interpretability of the judgments. 
%
%
Parmar \etal~\cite{parmar2017learning} assume that all the sub-action clips of the performance contribute equally towards the final score. In addition to back propagating gradients for the final loss, they also back propagate gradients for the individual sub-actions.
However, considering equal contributions of all clips during training is not realistic as different segments of an action have different contributions to the final score. 
%
In \cite{li2018end,xiang2018s3d}, authors compute the contributions of different segments of a performance towards the final scoring. 
This is achieved by learning separate regression models for a segment towards the final score. 
These regression models then predict the sub-action scores of the test videos. The rank correlation between the predicted sub-action scores and the final scores define the level of contribution of the sub-actions towards the final score. For eg. in diving, the splashes are seen to have the maximum contribution towards the scoring. Validation of the sub-actions scores of the individual videos has not been carried out.
%
%
 
A more meaningful solution to bring more objectivity and interpretability to an action assessment system is to consider a set of reference action videos as a pivot to evaluate a new action video. Such reference videos are expert's videos, which have the action performed with high precision. Thus, the problem of action quality assessment can be transformed into the problem of comparing a given action video with a reference video. 
%
However, one practical challenge is the lack of availability of a large number of reference videos to learn a regression model. Therefore, instead of learning how to compare a given action with an expert's action, we simply learn how to compare two actions. In our work we first train a metric learning module that can compare any two performances and predict a binary similarity label, and then learn a regression module that can use the learned metric to compare it with an expert's performance. 
We introduce a novel deep learning-based approach for AQA in which the performances are compared with a reference performance to determine the score. The proposed scoring system works in two phases (Fig.~\ref{fig:cover}):
\begin{itemize}
    \item A Deep Metric Learning (DML)~\cite{yi2014deep} module learns the similarity metric between two performances using the Siamese-LSTM network. The target label for this module is a binary label indicating similar/dis-similar actions.   
    Training this module requires inputs as pairs of videos. 
    This has the advantage that even smaller AQA datasets are able to generate a large number of pairs for training. 
    \item  In the second phase, a score prediction module is developed that uses the learned similarity metric and estimates the score of a video based on its resemblance to the reference video. 
    The input for training this module is a pair $-$ 
    an action video and an expert's video performing the same action. 

\end{itemize}
%
%





Additionally, we attempt to find the segment (sub-activity) level similarity using the learned DML network and identify how different segments contribute to the final score of the performance. This helps to make the judgement more  interpretable in terms of how a particular action in a video segment is performed compared to an expert performance. The sub-action (segment) level feedback can be verified by experts and can be used as a ground truth for future works. 
 
The key contributions of this work are as follows:
\begin{enumerate}
    \item We introduce a novel approach for automated AQA in which the predicted score depends on a relative comparison between the action performances in the test video and a  reference expert video.
    \item We train a DML network to learn the similarity metric between two action sequences based on the difference in the ground truth scores. Subsequently, a scoring network then regresses the concatenated representations (from the DML module) of the test video and a reference video to the final score.
    \item \textit{Feedback Proposal - }The learned DML network is also utilized to provide an unsupervised way of finding segment (sub-action) level contributions towards final scoring thereby making the scoring more interpretable.
    \item Through experiments we demonstrate the superiority of the proposed framework over the state-of-the-art solutions for AQA on publicly available Olympics diving and gymnastic vaults datasets. 
\end{enumerate} 



The paper is organized as follows : Section~\ref{sec:literature} discusses the existing literature towards human action assessment and action scoring. Section~\ref{sec:scoring_model} explains our scoring model architecture. Section~\ref{sec:feedback} describes a technique to generate sub-action level feedback for the performers. Section~\ref{sec:experiments} includes the experimental results using the proposed approach and discusses the performance of the proposed model. Finally the paper is concluded in Section~\ref{sec:conclusion}.

\section{Related Work}
\label{sec:literature}
The field of human action assessment is relatively new and there are only few works that have addressed the problem. Few early works \cite{jain2018detecting,jain2016framework,pervse2007automatic,jug2003trajectory,gordon1995automated} in the domain were hand crafted for specific actions and could not be generalised to different types of actions. Recently, there has been  development of more generic assessment frameworks, that can broadly be divided into two categories: 1) Action Quality Assessment (AQA) and 2) Skill Assessment. 

\subsection{Action Quality Assessment (AQA): }
Quality scoring for human actions has been posed as a supervised regression task, where  a model is learned to map the human action features to the ground-truth scores annotated by a domain expert. 

Pirsiavash \etal~\cite{pirsiavash2014assessing} utilized the low frequency DCT/DFT components of the pose features as input to the Support Vector Regressor (SVR) to map to the final action quality score. The pose features were estimated using Flexible Parts Model \cite{yang2011articulated} for each frame individually. Use of low frequencies filtered out high frequency noise due to pose estimation errors. Temporal resolution of features was improved with window-based extraction of frequency components. 

In addition to the holistic scores prediction, the authors propose a feedback scheme in terms of giving directions where the body parts should move to maximize the score. This is accomplished by differentiating the scoring function with respect to joint location. The maximum value of the gradient of the score with respect to the location of each joint indicates the direction that the performer must move to achieve the largest improvement in the score. The performance of the proposed assessment scheme was evaluated on the \textit{MIT Olympics dataset} that consisted of two action categories: Olympics diving and figure-skating sports. 

Venkataraman \etal~\cite{venkataraman2015dynamical} calculated the approximate entropy features using the estimated pose for each frame and concatenated them to get a high-dimensional feature. These features were seen to better encode the dynamical information compared to DCT and provide a better assessment over the MIT Olympics dataset. 

Pose features provide interpretable feedback such as directions in which the limbs should be moved to maximize scores. However, errors in pose estimation can negatively impact the scoring models. Moreover, pose-based representations are not capable of modeling objects used during an action (such as sports ball or tools), and do not consider physical outcomes, e.g. splashes in diving, which may be important features for some activities.

Visual spatio-temporal features such as Convolution 3D features (C3D) \cite{tran2015learning} and Pseudo 3D convolution features (P3D) \cite{qiu2017learning}, learned from 3D convolution neural networks capture both appearance and subtle motion cues in videos naturally. These features have been recently utilized to represent human actions for the purpose of scoring \cite{parmar2017learning}\cite{xiang2018s3d}\cite{li2018end}. Feature extraction has been done using pre-trained models \cite{parmar2017learning}\cite{li2018end} or learned in an end-to-end setup with a scoring objective \cite{xiang2018s3d}. 

Parmar and Morris~\cite{parmar2017learning} proposed three frameworks for human action quality assessment. These were: C3D-SVR, C3D-LSTM, and C3D-LSTM-SVR. The three frameworks differed in the way the features were aggregated and the regression was formulated. The authors proposed a new \textit{ UNLV sports dataset} which had twice the number of diving examples than the MIT dataset and also had gymnastic vault examples. The frameworks proved to be more efficient than the pose-based scoring works \cite{pirsiavash2014assessing, venkataraman2015dynamical}.

The authors~\cite{parmar2017learning} introduced a new training protocol called the \textit{incremental label training} to provide sub-action level feedback. It is expected that as an action advances in time, the score should build up (if the quality is good enough) or be penalized (if the quality is sub par). This intuition was formulated as a score getting accumulated as a non-decreasing function throughout an action. Thus a video is divided into equal sized non-overlapping clips such that each clip is assumed to have an equal contribution to the final score. Incremental-label training is used to guide the LSTM during the training phase to generate the final score along with intermediate score outputs (i.e. back-propagation occurs after each clip). The temporal score evolution as it changes through the LSTM structure is utilized to identify both ``good" and ``poor" segments of an action. However, training a network with an assumption that the clips have an equal contribution to the final score is not realistic. 

Xiang \etal~\cite{xiang2018s3d} proposed to divide a video into action specific semantic segments and fused the segment-averaged P3D features to learn the final score. The segment-level features were learned by fine-tuning the P3D network to the score regression task.  Li \etal~\cite{li2018end} divided a video sample into 9 clips and used 9 different C3D networks dedicated to different stages of Diving. The clip-level features were concatenated and fed to a dense network to produce a final AQA score. The network was optimized using the ranking loss along with the L2 loss. The ranking loss ensured the right rank order of the predicted scores. 

In both the aforementioned works \cite{xiang2018s3d,li2018end}, authors  contribute  towards  finding  the  contributions  of different segments of a performance towards the final scoring. Different  sub-action segments of all the training videos are separately regressed to  the final scores. The  rank  correlation  between  the predicted sub-action scores and the final scores define the level of contribution of the sub-actions towards the final score. For e.g.,  in diving, the splashes  are  seen  to  have  the  maximum contribution  towards  the  scoring. 

More recently, Xu \etal~\cite{xu2019learning} addressed the problem of scoring figure skating. Figure skating videos are long and last for few minutes unlike vaults and diving that last for only few seconds. The assessment of such videos requires evaluation of temporal segments as well as holistic performance. The authors proposed a deep architecture with 2 complementary components: Self-Attentive LSTM and Multi-scale Convolutional Skip LSTM, which learn the local and global sequential information respectively from a video. They introduced \textit{FisV}, a new large scale figure skating video dataset for evaluating the proposed scheme. The attention weight matrix computed using the self attentive mechanism for a specific video signifies clip-level significance. A clip that has high weight in at least one row of the attention weight matrix shows an important technical movement and is otherwise considered insignificant. The multi-scale LSTM employs several parallel 1D convolution layers with different kernel sizes unlike C3D or P3D that have a fixed kernel size. A kernel with small size extracts the visual representation of action patterns lasting seconds in the videos. A kernel of large size tries to model the global information of the videos.


In this work, we develop an  action assessment system which can provide segment-level feedback. For learning a scoring model we consider a set of reference videos, which we call as \textit{expert videos}, i.e. the highest rated performers for the same type of action. We transform the problem of action assessment into the problem of 
(i) Generating a video representation which incorporates comparison with an expert video of the same type of action. 
(ii) Mapping the learned representation to the aggregate score using a regression model. 
The same network is also used to do a segment level comparison with a corresponding expert video segment and generate scores which depict assessment of the sub-action performed in that segment. The proposed system thus provides a more interpretable assessment.

Some works have addressed classifying performances into skill levels. 
Zia \etal~\cite{zia2015automated} extract spatio-temporal interest points (STIP’s) in the frequency domain to classify the performances into amateur, intermediate or expert skill levels. Doughty \etal~\cite{doughty2018s} learned convolutional features with ranking loss objective function to evaluate surgical, drawing, chopstick use and dough rolling skills. In their next work \cite{doughty2018pros}, they use temporal attention LSTMs to learn weighted contributions of different segments of the video to the final skill levels. The benefit of using Siamese networks for the task of learning the relative skills of performances has encouraged us to adopt a similar approach to the task of action quality assessment and provide interpretable feedback along with the scores. 


In this work, we propose a new action quality assessment approach that transforms the problem of assessment into the problem of comparing a given action video with a reference video. Thus we require a model that can learn the similarity between two action sequences. This problem relates closely with the problem of estimating  similarity of two signature samples. A Siamese network \cite{chopra2005learning} is widely used as a deep metric learning-based (DML) approach \cite{yideep} to learn the similarity between two sequences. The DML approach has been used for human action recognition in \cite{yucer20183d} and has shown promising results. While recognition is an inter-class discrimination problem, it remains to be explored if DMLs can   learn differences between very similar sequences of the same class. The level of resemblance can be translated into scores pertaining to assessment of actions. We discuss our model in Section~\ref{sec:scoring_model}.

\subsection{Action Quality Assessment Datasets}
The past works have contributed Olympics dataset for 3 actions: Diving, Gymnastic Vaults and Figure Skating. 

\textit{Diving Dataset:} The MIT-Diving dataset \cite{pirsiavash2014assessing} consists of 159 videos taken from 2012 Olympic men’s 10-meter platform prelims round. The UNLV-diving dataset \cite{parmar2017learning} is an extension to this dataset which includes semi-final and final round videos totaling to 370 videos, each having around 150 frames.  A diving score is determined by the product of execution score, based on judge's assessment of the quality of diving, multiplied by the diving difficulty score, which is a fixed value based on diving type. The scores vary between 0 (the worst) and 100 (the best).

More recently Parmar \etal~\cite{parmar2019and} proposed a Multitask AQA Dataset with 1412 samples of diving videos including 10m Platform as well as 3m Springboard with both male and female athletes, individual or pairs of synchronized divers, and different views. The AQA score, dive type and the commentary for each sample has been included in the dataset.

\textit{Vault Dataset}: The UNLV-Vault Dataset \cite{parmar2017learning} dataset includes 176 videos. These videos are relatively short with an average length of about 75 frames.  A vault score is determined by the sum
of the execution score and the difficulty score. The score ranges from 0 (the worst) to 20 (the Best). 

\textit{Figure Skating:} Pirsiavash \etal~\cite{pirsiavash2014assessing} proposed the MIT-skate dataset has 150 videos with 24 frames per second. On an average, figure skating samples are 2.5 minutes long and with continuous view variation during a performance. The judge’s score ranges between 0 (worst) and 100 (best). Xu \etal~\cite{xu2019learning} proposed the Fis-V dataset which consists of 3 times more videos than the MIT-skate dataset.

In this work we consider two action types - diving and gymnastic vaults. We do not consider figure skating because it is characterized by large variations of routines attempted by performers and does not have a standard template and therefore not suitable for assessment using comparison. 
%
\begin{figure*}[h!]
    \centering
    \includegraphics{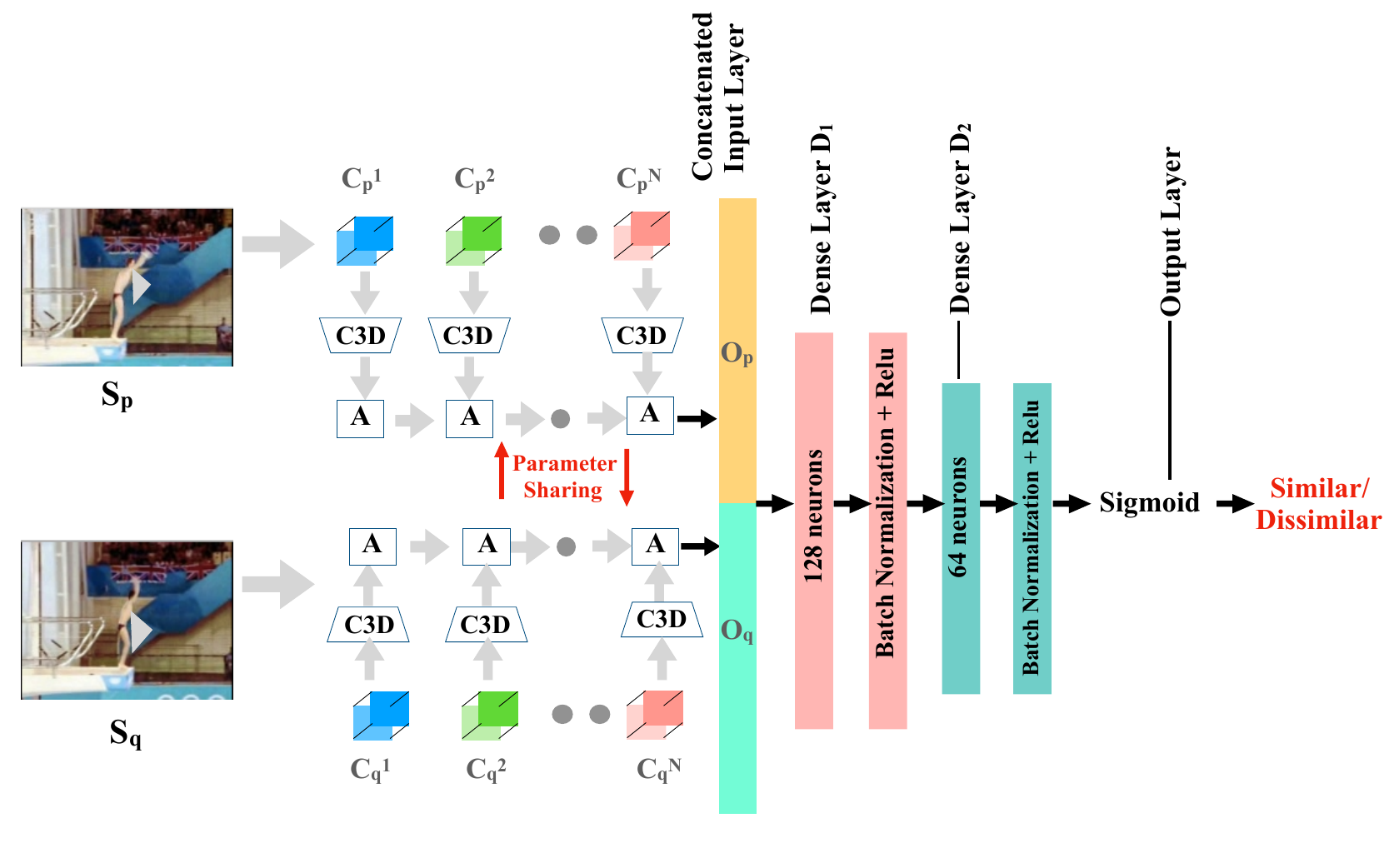}
    \caption{Siamese Architecture for Deep Metric Learning }
    \label{fig:siamese}
\end{figure*}

\section{Proposed Scoring Model}
\label{sec:scoring_model}
In this work, we address the problem of human action scoring as the problem of comparing a given action video with a reference video depicting a high rated performance. Our proposed assessment system is a two-phase system: 1) A \textit{Deep Metric Learning Module} which utilizes a Siamese Network to learn a similarity metric between any two action sequences. 2) A \textit{Score Estimation Module} which utilizes the learned Siamese Network and uses regression to estimate the score of a video based on its similarity to the expert execution of the same action.

We claim that learning a similarity metric before training a scoring regression model offers an advantage because even small training datasets can give a large number of pairs to train
the DML network. A pair need not include an expert sequence. We explain these steps in detail.

\subsection{Deep Metric Learning Module}

The Deep Metric Learning (DML) model is trained to learn the similarity between any two videos in a pair. The pair of sequences to be used as training input are sampled from the set of all videos.
Fig.~\ref{fig:siamese} shows the structure of the proposed metric learning module. This module takes two video sequences $V_p$ and $V_q$ as input and produces similar/dissimilar $(1/0)$ label as output. Similar to \cite{parmar2017learning}, the clip-level C3D FC-6 layer features \cite{tran2015learning} are fed to an LSTM layer to generate a video-level description. 
The target labels to be learned for the model are binary valued $1/0$, which classify the pairs to be either similar or dissimilar. The target label for each pair is decided based on the difference in their respective scores assigned by the judges. If the difference in the scores is less than a threshold $th$, the pairs are considered as similar (label 1), and otherwise dissimilar. The threshold is decided based on the type of activity being assessed. We would discuss the setting of this threshold in the experiments section.

The C3D model provides a more compact video representation than frame-level CNN. Since an entire clip is processed in one go, this results in fewer processing steps. For eg. a video with $145$ frames can be represented as a sequence of $9$ clips each containing $16$ frames (input size required for the C3D network). 
Each of the two LSTM networks takes one sequence as input and  produces individually two output vectors $O_p \in R^M$ and $O_q \in R^M$. These vectors are then concatenated and fed into 2 fully connected dense layers, $D_1$ and $D_2$. The output of the dense layer is passed to one fully connected Sigmoid layers to map the combined features of the two videos to a binary  similar/non-similar classification output.

The effectiveness of the Deep Metric Learning module is in terms of how well it transforms the inputs to a feature space such that the videos with similar scores are closer in the feature space and those with a large difference in scores are moved farther apart.

\subsection{Score Estimation Module using Expert bias}

Our proposal for action quality assessment is to compare a video with an expert (highest rated) video and map the result of comparison to its final score. This is different from how the traditional scoring models have been working where a video representation is directly regressed to the scores.

An action can be performed in various ways. For eg. diving can be performed in forms like forward somersaults, inward somersaults, backward somersaults, etc. In order to do scoring in our model, each video is compared to an expert's performance of the same  type of action.  
Further, we can have one or more experts for an action type while training the scoring model. 

The scoring model utilizes the DML Siamese model, which is
a subset of the architecture shown in Fig.~\ref{fig:siamese}. The model takes input as a pair with an expert and a non-expert video depicting the same type of action. The last sigmoid layer is replaced with a fully connected (FC) layer, so that the Siamese network model originally trained for classifying a pair of sequences as similar or dissimilar can now be used for the regression task and estimate the relative score of the video. 
The LSTM-Siamese weights are learned while training the DML module. These weights are fixed during the learning of the score estimation module. Instead, only the weights for the final connected layer which performs regression are learned. 

To summarize, the DML module is trained with action pairs that include any action sequence $V_p$ and $V_q$ as input and the match or no-match labels as output, while the score estimation module is trained with the input as a pair of videos with $E_{t}$ being an expert video of type $t$ and $V_{qt}$ is the $q^{\text{th}}$ training video of the same action type $t$ and the output is the ground truth score of the sequence $V_{qt}$. 

\begin{figure}[t!]
    \centering
    \includegraphics[scale=0.5]{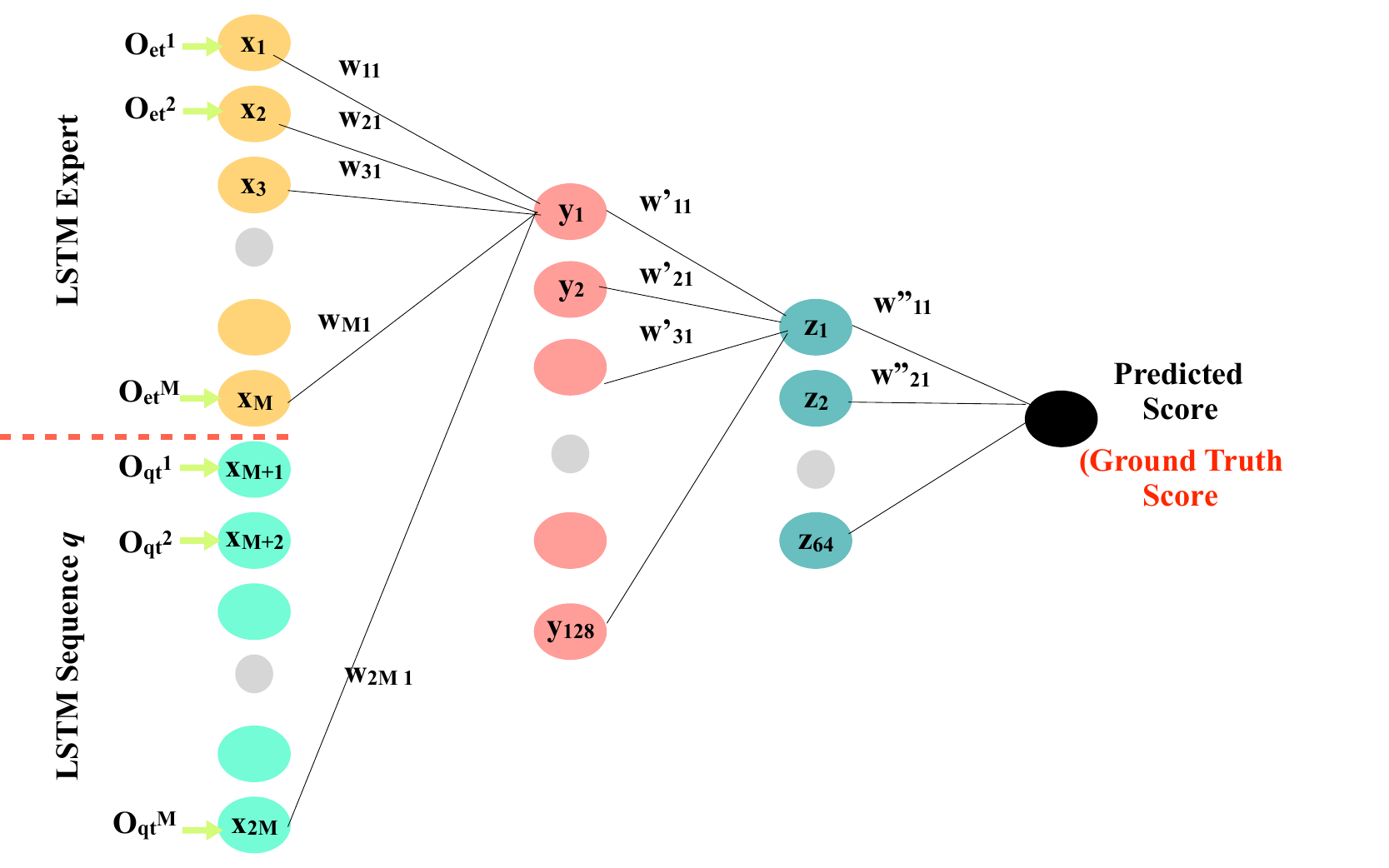}
    \caption{Score Estimation Relative to Expert}
    \label{fig:score}
\end{figure}

\begin{figure*}[t!]
    \centering
    \includegraphics[scale=0.9]{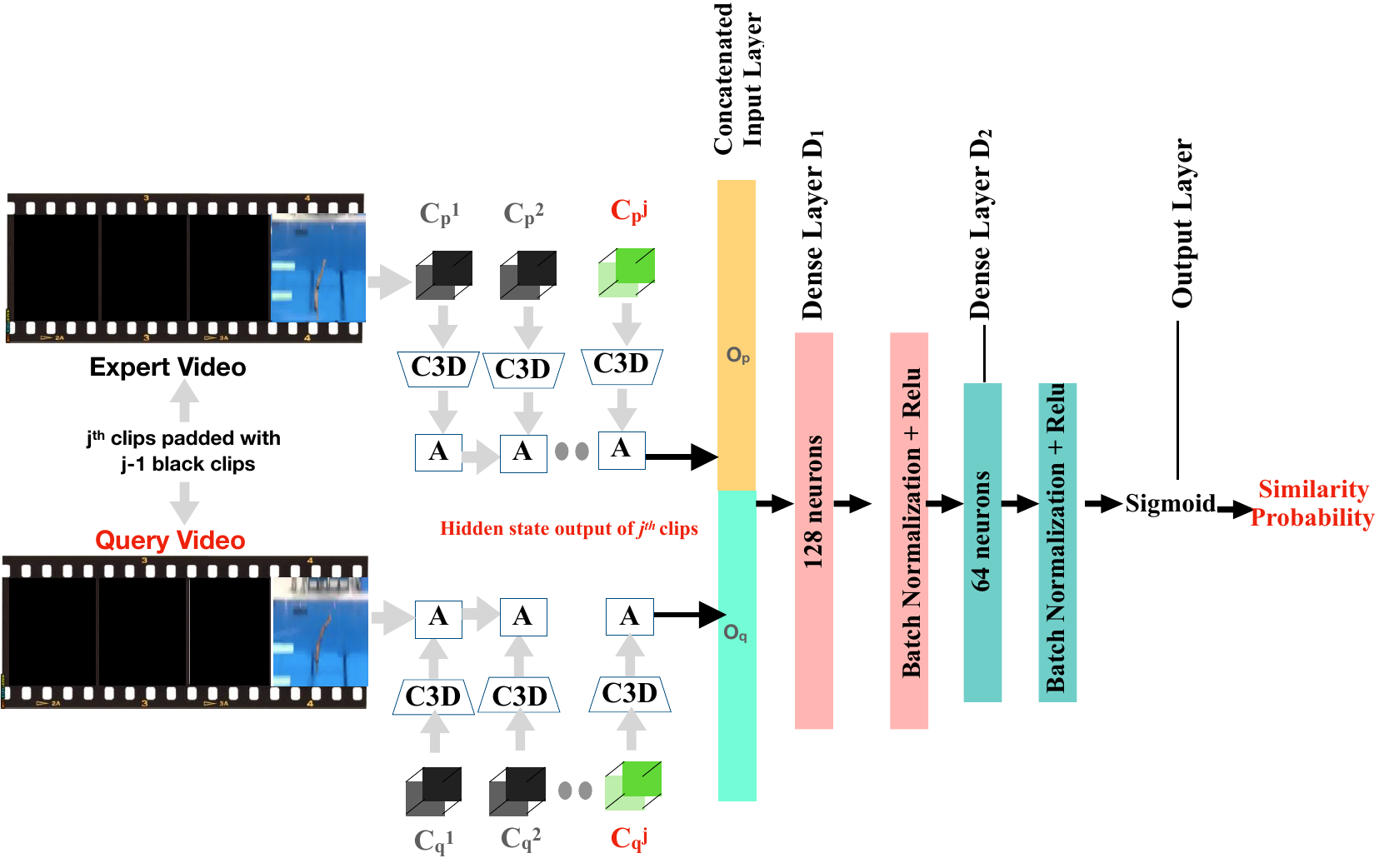}
    \caption{Clip level similarity Estimation for evaluating individual clip contribution}
    \label{fig:clip_similarity}
\end{figure*}

\textit{Expert performance as bias to the network}
The Siamese embedding of the expert acts as a fixed but dive type dependent bias for the regression network, because the expert is fixed for multiple videos of the same action type. 
Here we show that 
the gradient of the loss used to update the weights $w''$ (Fig.~\ref{fig:score}) of the last layer contains a contribution of expert's bias which is different for different dive types. 
Let the LSTM embedding of the expert videos $E_{t}$ for action type $t$ in the pair be $O_{et} \in R^M$ and that of the $q^{th}$ training video be $O_{qt} \in R^M$. Let the concatenation of LSTM output of the two vectors be represented as vector $X \in R^{2M}$ and likewise the outputs of the dense layers $D_1$ and $D_2$ be represented as $Y \in R^{128}$ and $Z \in R^{64}$. 

The weights of the dense layers ($w$ from $X$ to $Y$ and $w'$ from $Y$ to $Z$) are fixed and are previously learned during DML training. During score estimation only the weights from layer $Z$ to the final regression layer are learned with the target of the ground truth score provided by judges. The scores are estimated following these steps: 
\begin{enumerate}
    \item The vector $Y$ for action type $t$ and $q^{th}$ training video of the same type is computed as
    \[
y_{qi}^t = \sum_{j=1}^{M} w_{ji}x_j^t + \sum_{j=M+1}^{2M} w_{ji}x_{qj}^t 
\]
where the first term is the contribution from the expert and is common for all training videos of one type. We ignore the bias to keep the calculations easy.

\item Similarly the vector $Z$ for action type $t$ and $q^{th}$ training video is computed as
\[
z_{qk}^t = \sum_{i=1}^{128} w'_{ik} ( \sum_{j=1}^{M} w_{ji}x_{j}^t + \sum_{j=M+1}^{2M} w_{ji}x_{qj}^t ) 
\]

\[
= \sum_{i=1}^{128} w'_{ik} \sum_{j=1}^{M} w_{ji}x_j^t + \sum_{i=1}^{128} w'_{ik}\sum_{j=M+1}^{2M} w_{ji}x_{qj}^t  
\]

the first term is again the contribution from the expert, common for all training videos of one type and the second term depends on the $q^{th}$ training video of type $t$. Let the first term be represented by $a_{ek}^t$ and the second term be $zx_{qk}^t$. Thus $z_{qk}^t = a_{ek}^t + zx_{qk}^t $

\item The final score is thus computed as 

\[
S^{'t}_{q} = \sum_{k=1}^{64} w'' ( a_{ek}^t + zx_{qk}^t) 
\]

\end{enumerate}

Let the ground truth score of the $q^{th}$ video be $S_q$. The total loss over all training examples for all types is:

\[
Loss = \sum_{t=1}^{\#action\_types} \sum_{q \in t} (S_q - \sum_{k=1}^{64} w''_k ( a_{ek}^t + zx_{qk}^t))^2
\]

Thus, while learning to estimate the scores of the videos, experts of different action types contribute in different ways during relative score prediction. To make the settings more general, we can consider the bias as a contribution by any kind of performance, e.g. it could be expert, intermediate or worst performer.   

\subsection{Module Training}
When training the DML module, there is a label imbalance problem because the number of non-matching pairs is much larger than the matching ones. To keep the training set balanced, we increase the match pairs by using augmentation techniques like video zoom, brightness alteration, temporal augmentation by dropping some random frames, etc. 
Training of the score estimation module follows the training of DML module. 
The score estimation module is trained using only the original training dataset, without using any augmentation.  

\section{Feedback Proposals}
\label{sec:feedback}

 Parmar \etal~\cite{parmar2017learning} proposed an incremental label training where it was expected that the temporal score evolution as it changes through the LSTM should be monotonically increasing. The ``good" and ``poor" components of an action are identified when the evolution violates monotonicity and the errors are expected to result in a loss of score. Following this intuition, the predicted sub-scores from the incremental training approach are analysed and the faulty clips are identified. 

Events like Olympics Diving and Gymnastic vaults are annotated only by a final score provided by the judge. In order to make the assessments more interpretable and estimate how different parts of the videos contribute to the final score, we compute the similarity of each clip of a test video to the corresponding clip of the expert video. 
To compute the similarity of the $j^{\text{th}}$ clips of the two videos, we first create the corresponding trimmed sequences which retain the  $j^{\text{th}}$ clips, but remove the succeeding frames and zero out all the preceding frames. Feeding the trimmed video as input to the LSTM provides the  final hidden states at the end of the $j^{\text{th}}$ clip. LSTM hidden states for the test and the expert video are concatenated and fed to the dense layers of the Siamese network. (Fig.~\ref{fig:clip_similarity}) 
The final Sigmoid layer gives the probability of similarity of the two clips. A higher probability implies that the actions corresponding to the $j^{\text{th}}$ clips of the two videos were performed in a similar way. 


\section{Experiments} 
\label{sec:experiments}
We compare our relative scoring approach to the baseline absolute scoring techniques in this section.  

\subsection{Dataset Splits}
The experiments are based on the Olympics UNLV Diving Dataset and the Gymnastic Vaults Dataset\cite{parmar2017learning}. For the diving dataset, $10$ training splits are used to compare the performance of our model to the baseline models. For training the DML module we choose the top $3$ most frequent dive types from the UNLV dataset while for training the scoring module we choose the top $6$ most frequent dives.
By examining the scoring performance on the dive types which were not seen during training, we can comment on the generalizability of DML training. 

For training the DML module we use $10$ splits of the training set, each with $130$ videos while for training the score estimation module we use $10$ splits, each with $174$ training videos. We used $90$ test videos which were not included during DML training. 
 
 We assign the match/mismatch targets as follows.    
    We first compute the sum of the scores awarded by the three judges. The score awarded by a judge is in the range 0 to 10 and has a deviation of $\pm$0.5.  This implies that while judging the same performance a judge may not give exactly the same score every time, but may give a slightly different score with a tolerance of up to $\pm$0.5. Thus the sum of the scores given by the three judges may have a total deviation $\pm$1.5. 
    Since the overall score is computed as the product of the judge's score with the difficulty level, which is 3$\pm$0.3, the total deviation in the overall score is 
    $\pm 0.5\times0.33\approx \pm 5$. 
    We consider two videos to be similar if their overall scores do not differ by more than $\pm$5.
    The positive pairs to be used for training the Siamese Metric Learning network are identified as the ones with the two videos having a score difference of less than 5. 
    
    
    Based on this criteria, there are approximately $1300$ positive pairs and $1900$ negative pairs in the splits.  We augment the training videos by applying augmentation techniques like variations in brightness, zooming effect, masking of the background and histogram equalization to give us about $26000$ pairs. 
    
    The UNLV vaults dataset\cite{parmar2017learning} contains 176 videos with 3 vault types out of which $138$ vaults are performed in tuck position while the rest are free and twist vault types. We consider only the $138$ tuck vaults for our experiment because the other vault types have inadequate samples for training and testing. Similar to the diving setup, we evaluate the scoring model for $5$ train-test split, where the training set comprises $98$ videos and test set comprises $40$ videos. 
    We train only the scoring module using the vault dataset and use the same DML module that was trained on the diving dataset. 
    The vaults action is considered to show the generalizability of the DML model that has been trained for diving action. 
    
    \subsection{Performance evaluation metrics }
    We evaluate the performance of the models using two metrics:
    
    \begin{enumerate}
    \item Rank correlation between the predicted and the ground truth scores : Spearman rank correlation $\rho$ \cite{pirsiavash2014assessing,parmar2017learning} is used to measure performance. Higher $\rho$ signifies better rank correlation between the true and predicted scores. This metric allows for non-linear relationship, however, it does not explicitly emphasizes the true score value, but the relative ranking.
    \item Mean Square Error between the predicted and the ground truth scores: This metric emphasizes that the predicted score should be as close as possible to the score.
    \end{enumerate}

    \subsection{Baseline Works}
    We compare our works with three \cite{pirsiavash2014assessing,parmar2017learning,xiang2018s3d} baseline techniques for action score estimation.  While \cite{pirsiavash2014assessing} utilizes pose-based features, the other two techniques \cite{parmar2017learning,xiang2018s3d} use spatio-temporal features. Also, \cite{pirsiavash2014assessing,parmar2017learning} are segmentation free approaches to score estimation i.e. the information about boundaries of the semantic segments is not available, while  \cite{xiang2018s3d} requires the knowledge of the semantic boundaries. Both variants of the C3D+LSTM models proposed by Parmar \etal~\cite{parmar2017learning} \textit{i.e.} final score labeling (F) and incremental training (I) are used in comparison. We use the code provided by the authors to evaluate their performance on our dataset splits. 
    

\begin{table*}[t!]
\centering
\caption{Comparison of Rank Correlation; C:C3D, L:LSTM, F: Final, I: Incremental, S: SVR}
\label{tab1}
\footnotesize{}
\setlength\tabcolsep{4pt}
\smallskip


\begin{tabular}{|c|c|c|c|c|c|c|c|c|c|c|c|}
\hline
\multicolumn{1}{|l|}{} & \multicolumn{7}{c|}{\textbf{Baseline Works}} & \multirow{3}{*}{\textbf{\begin{tabular}[c]{@{}c@{}}Dive Specific \\ Regression\\ models\end{tabular}}} & \multicolumn{3}{c|}{\textbf{Ours}} \\ \cline{1-8} \cline{10-12} 
\multicolumn{1}{|l|}{\multirow{2}{*}{}} & \multicolumn{2}{c|}{\multirow{2}{*}{\textbf{Pirsiavash \etal~\cite{pirsiavash2014assessing}}}} & \multicolumn{4}{c|}{\multirow{2}{*}{\textbf{Parmar \etal~\cite{parmar2017learning}}}} & \multirow{2}{*}{\textbf{Xiang \etal~\cite{xiang2018s3d}}} &  & \multicolumn{3}{c|}{\multirow{2}{*}{\textbf{Scoring with Bias}}} \\
\multicolumn{1}{|l|}{} & \multicolumn{2}{c|}{} & \multicolumn{4}{c|}{} &  &  & \multicolumn{3}{c|}{} \\ \hline
\textbf{Splits} & \textbf{\begin{tabular}[c]{@{}c@{}}Pose \\ + DCT\end{tabular}} & \textbf{\begin{tabular}[c]{@{}c@{}}Pose \\ + DFT\end{tabular}} & \textbf{\begin{tabular}[c]{@{}c@{}}C + L \\ (F)\end{tabular}} & \textbf{\begin{tabular}[c]{@{}c@{}}C + L\\  (I)\end{tabular}} & \textbf{\begin{tabular}[c]{@{}c@{}}C + L\\ + S(F)\end{tabular}} & \textbf{\begin{tabular}[c]{@{}c@{}}C + L\\ + S(I)\end{tabular}} & \textbf{S3D} & \textbf{\begin{tabular}[c]{@{}c@{}}C + L\\ (F)\end{tabular}} & \textbf{\begin{tabular}[c]{@{}c@{}}Constant\\ Expert \\ Bias\end{tabular}} & \textbf{\begin{tabular}[c]{@{}c@{}}Worst \\ performer\\ bias per dive\end{tabular}} & \textbf{\begin{tabular}[c]{@{}c@{}}Best\\ performer\\ bias per dive\end{tabular}} \\ \hline
1 & 0.46 & 0.42 & 0.65 & 0.6 & 0.62 & 0.46 & 0.52 & 0.69 & 0.67 & 0.69 & 0.71 \\ \hline
2 & 0.3 & 0.24 & 0.55 & 0.49 & 0.54 & 0.42 & 0.39 & 0.58 & 0.54 & 0.53 & 0.65 \\ \hline
3 & 0.53 & 0.42 & 0.63 & 0.62 & 0.65 & 0.5 & 0.39 & 0.71 & 0.69 & 0.7 & 0.71 \\ \hline
4 & 0.52 & 0.41 & 0.66 & 0.57 & 0.58 & 0.42 & 0.49 & 0.59 & 0.7 & 0.69 & 0.8 \\ \hline
5 & 0.43 & 0.4 & 0.68 & 0.58 & 0.61 & 0.67 & 0.42 & 0.68 & 0.62 & 0.66 & 0.73 \\ \hline
6 & 0.41 & 0.44 & 0.67 & 0.47 & 0.63 & 0.56 & 0.34 & 0.69 & 0.62 & 0.56 & 0.67 \\ \hline
7 & 0.49 & 0.56 & 0.65 & 0.49 & 0.58 & 0.49 & 0.58 & 0.68 & 0.64 & 0.7 & 0.72 \\ \hline
8 & 0.35 & 0.38 & 0.58 & 0.58 & 0.62 & 0.52 & 0.55 & 0.63 & 0.46 & 0.52 & 0.63 \\ \hline
9 & 0.49 & 0.52 & 0.66 & 0.63 & 0.66 & 0.59 & 0.47 & 0.65 & 0.58 & 0.72 & 0.65 \\ \hline
10 & 0.33 & 0.42 & 0.67 & 0.58 & 0.63 & 0.52 & 0.6 & 0.71 & 0.53 & 0.66 & 0.65 \\ \hline
\textbf{Average} & 0.43 & 0.42 & 0.64 & 0.56 & 0.61 & 0.51 & 0.47 & 0.66 & 0.6 & 0.64 & \textbf{0.69} \\ \hline
\end{tabular}
\end{table*}

\begin{table*}[t!]
\centering
\caption{Comparison of Mean Square Error; C:C3D, L:LSTM, F: Final, I: Incremental, S: SVR}
\label{tab2}
\footnotesize{}
\setlength\tabcolsep{4pt}
\smallskip
\begin{tabular}{|c|c|c|c|c|l|c|c|c|c|c|c|}
\hline
\multicolumn{1}{|l|}{} & \multicolumn{7}{c|}{\textbf{Baseline Works}} & \multirow{3}{*}{\textbf{\begin{tabular}[c]{@{}c@{}}Dive Specific\\ Regression\\ models\end{tabular}}} & \multicolumn{3}{c|}{\textbf{Ours}} \\ \cline{1-8} \cline{10-12} 
\multicolumn{1}{|l|}{\multirow{2}{*}{}} & \multicolumn{2}{c|}{\multirow{2}{*}{\textbf{Pirsiavash \etal~\cite{pirsiavash2014assessing}}}} & \multicolumn{4}{c|}{\multirow{2}{*}{\textbf{Parmar \etal~\cite{parmar2017learning}}}} & \multirow{2}{*}{\textbf{Xiang \etal~\cite{xiang2018s3d}}} &  & \multicolumn{3}{c|}{\multirow{2}{*}{\textbf{Scoring with Bias}}} \\
\multicolumn{1}{|l|}{} & \multicolumn{2}{c|}{} & \multicolumn{4}{c|}{} &  &  & \multicolumn{3}{c|}{} \\ \hline
\textbf{Splits} & \textbf{\begin{tabular}[c]{@{}c@{}}Pose\\  + DCT\end{tabular}} & \textbf{\begin{tabular}[c]{@{}c@{}}Pose \\  + DFT\end{tabular}} & \textbf{C+L(F)} & \textbf{C+L(I)} & \textbf{\begin{tabular}[c]{@{}l@{}}C + L\\ +S (F)\end{tabular}} & \multicolumn{1}{l|}{\textbf{\begin{tabular}[c]{@{}l@{}}C + L\\ +S (I)\end{tabular}}} & \textbf{S3D} & \textbf{\begin{tabular}[c]{@{}c@{}}C + L\\ (F)\end{tabular}} & \textbf{\begin{tabular}[c]{@{}c@{}}Constant\\ Expert \\ Bias\end{tabular}} & \textbf{\begin{tabular}[c]{@{}c@{}}Dive \\ Specific\\ Worst\\ Performer\\ Bias\end{tabular}} & \textbf{\begin{tabular}[c]{@{}c@{}}Dive\\  Specific\\ Expert\\ Performer\\ Bias\end{tabular}} \\ \hline
1 & 153.59 & 160.25 & 68.37 & 245.59 & 95.09 & 232.91 & 286.84 & 81.2 & 89.96 & 88.81 & 64.67 \\ \hline
2 & 212.12 & 228.48 & 107.81 & 252.03 & 108.04 & 226.36 & 192.4 & 108.26 & 114.5 & 98.54 & 101.31 \\ \hline
3 & 197.83 & 185.37 & 122.88 & 312.31 & 116.47 & 289.27 & 208.94 & 102.92 & 77.6 & 92.64 & 77.95 \\ \hline
4 & 214.02 & 228.15 & 104.11 & 344.2 & 124.25 & 383.6 & 262.73 & 133.27 & 100.46 & 92.96 & 72.43 \\ \hline
5 & 206.18 & 198.42 & 73.41 & 276.64 & 88.87 & 264.02 & 215.04 & 96.04 & 105.53 & 84.19 & 89.18 \\ \hline
6 & 184.56 & 169.37 & 84.11 & 249.26 & 108.19 & 234.83 & 208.62 & 72.43 & 97.18 & 97.36 & 85.08 \\ \hline
7 & 154.4 & 136.39 & 99.99 & 396.42 & 101.96 & 386.94 & 166.47 & 111.08 & 112.16 & 89.88 & 102.7 \\ \hline
8 & 153.96 & 158.76 & 106.34 & 223.33 & 102.89 & 225.67 & 160.34 & 110.68 & 146.15 & 137.58 & 110.17 \\ \hline
9 & 178.6 & 179.02 & 108.73 & 264.53 & 94.48 & 258.98 & 195.72 & 119.14 & 118.01 & 74.94 & 95.26 \\ \hline
10 & 231.25 & 213.58 & 74 & 327.81 & 113.33 & 303.89 & 231.18 & 80.81 & 110.17 & 90.32 & 87.2 \\ \hline
\textbf{Average} & 188.65 & 185.78 & 94.97 & 289.21 & 105.35 & 280.65 & 212.83 & 101.58 & 107.18 & 94.72 & \textbf{88.59} \\ \hline
\end{tabular}
\end{table*}

\begin{table}[]
\caption{Expert and Worst performer details for Diving}
\label{e_w_dive}
\begin{tabular}{|c|c|c|c|c|}
\hline
\textbf{\begin{tabular}[c]{@{}c@{}}Dive \\ Type\end{tabular}} & \textbf{\begin{tabular}[c]{@{}c@{}}Expert \\ Score\end{tabular}} & \textbf{\begin{tabular}[c]{@{}c@{}}Count of \\ Expert Performers\end{tabular}} & \textbf{\begin{tabular}[c]{@{}c@{}}Worst\\  Score\end{tabular}} & \textbf{\begin{tabular}[c]{@{}c@{}}Count of Worst\\ Performer\end{tabular}} \\ \hline
1                                                             & 92.8                                                             & 1                                                                              & 51.2                                                            & 1                                                                           \\ \hline
2                                                             & 102.6                                                            & 2                                                                              & 21.6                                                            & 1                                                                           \\ \hline
3                                                             & 94.05                                                            & 4                                                                              & 34.65                                                           & 1                                                                           \\ \hline
4                                                             & 99.9                                                             & 2                                                                              & 66.6                                                            & 1                                                                           \\ \hline
5                                                             & 99.75                                                            & 1                                                                              & 36.3                                                            & 1                                                                           \\ \hline
6                                                             & 102.6                                                            & 1                                                                              & 64.8                                                            & 1                                                                           \\ \hline
\end{tabular}
\end{table}

\begin{table*}[]
\centering
\caption{Comparison of different techniques : Rank Correlation and Mean Square Error for individual Dive Types}
\label{tabb}
\footnotesize{}
\setlength\tabcolsep{4pt}
\smallskip
\begin{tabular}{|c|cc|c|c|c|c|c|c|c|l|l|c|c|c|c|c|}
\hline
                                                             & \multicolumn{14}{c|}{\textbf{Baseline Works}}                                                                                                                                                                                                                                                                                                                                                                                                                                                                                                                        & \multicolumn{2}{c|}{\multirow{2}{*}{\textbf{Ours}}}                                                    \\ \cline{1-15}
                                                             & \multicolumn{4}{c|}{\textbf{Pirsiavash \etal~\cite{pirsiavash2014assessing}}}                                                                                                                                & \multicolumn{8}{c|}{\textbf{Parmar \etal~\cite{parmar2017learning}}}                                                                                                                                                                                                                                                                                                                 & \multicolumn{2}{c|}{\textbf{Xiang \etal~\cite{xiang2018s3d}}} & \multicolumn{2}{c|}{}                                                                                  \\ \hline
\textbf{\begin{tabular}[c]{@{}c@{}}Dive\\ Type\end{tabular}} & \multicolumn{2}{c}{\textbf{\begin{tabular}[c]{@{}c@{}}Pose\\ + DCT\end{tabular}}} & \multicolumn{2}{c|}{\textbf{\begin{tabular}[c]{@{}c@{}}Pose \\ + DFT\end{tabular}}} & \multicolumn{2}{c|}{\textbf{\begin{tabular}[c]{@{}c@{}}C + L\\  (F)\end{tabular}}} & \multicolumn{2}{c|}{\textbf{\begin{tabular}[c]{@{}c@{}}C + L \\ (I)\end{tabular}}} & \multicolumn{2}{c|}{\textbf{\begin{tabular}[c]{@{}c@{}}C+L\\ +S (F)\end{tabular}}} & \multicolumn{2}{c|}{\textbf{\begin{tabular}[c]{@{}c@{}}C + L \\ + S(I)\end{tabular}}} & \multicolumn{2}{c|}{\textbf{S3D}}   & \multicolumn{2}{c|}{\textbf{\begin{tabular}[c]{@{}c@{}}Best\\ performer\\ bias per dive\end{tabular}}} \\ \hline
\textbf{}                                                    & \multicolumn{1}{c|}{\textbf{RC}}                  & \textbf{MSE}                  & \textbf{RC}                              & \textbf{MSE}                             & \textbf{RC}                             & \textbf{MSE}                             & \textbf{RC}                             & \textbf{MSE}                             & \textbf{RC}                             & \textbf{MSE}                             & \textbf{RC}                               & \textbf{MSE}                              & \textbf{RC}      & \textbf{MSE}     & \textbf{RC}                                       & \textbf{MSE}                                       \\ \hline
1                                                            & \multicolumn{1}{c|}{0.243}                        & 105.92                        & 0.23                                     & 123.48                                   & 0.62                                    & 53.03                                    & 0.64                                    & 61.73                                    & 0.53                                    & \textbf{41.17}                           & 0.49                                      & 104.39                                    & 0.05             & 96.58            & \textbf{0.67}                                     & 44.31                                              \\ \hline
2                                                            & \multicolumn{1}{c|}{0.55}                         & 227.98                        & 0.6                                      & 189.58                                   & 0.69                                    & 132.88                                   & 0.73                                    & 118.21                                   & 0.68                                    & 157.71                                   & 0.68                                      & 434.54                                    & 0.48             & 323.13           & \textbf{0.76}                                     & \textbf{105.62}                                    \\ \hline
3                                                            & \multicolumn{1}{c|}{0.36}                         & 208.55                        & 0.4                                      & 188.61                                   & \textbf{0.69}                           & 102.3                                    & 0.65                                    & 113.53                                   & 0.59                                    & 98.19                                    & 0.56                                      & 241.62                                    & 0.39             & 188.06           & \textbf{0.69}                                     & \textbf{95.43}                                     \\ \hline
4                                                            & \multicolumn{1}{c|}{0.12}                         & 139.84                        & 0.25                                     & 156.71                                   & 0.24                                    & 90.08                                    & 0.3                                     & 118.3                                    & 0.29                                    & 105.9                                    & 0.23                                      & 202.14                                    & 0.019            & 107.79           & \textbf{0.34}                                     & \textbf{81.99}                                     \\ \hline
5                                                            & \multicolumn{1}{c|}{0.34}                         & 351.72                        & 0.27                                     & 356.15                                   & 0.79                                    & \textbf{90.34}                           & 0.73                                    & 116.56                                   & \textbf{0.8}                            & 107.25                                   & 0.7                                       & 407.48                                    & 0.18             & 453.48           & 0.78                                              & 104.06                                             \\ \hline
6                                                            & \multicolumn{1}{c|}{0.37}                         & 97.89                         & 0.14                                     & 100.16                                   & 0.43                                    & 101.23                                   & 0.42                                    & 117.38                                   & 0.47                                    & 130.91                                   & 0.41                                      & 163.93                                    & 0.38             & 107.92           & \textbf{0.48}                                     & \textbf{97.16}                                     \\ \hline
\end{tabular}
\end{table*}

\begin{table}[ht!]
\centering
\caption{Study of Impact of Metric Learning on final scoring}
\label{fig:tab3}
\begin{tabular}{|c|c|c|c|}
\hline
\textbf{Model}                             & \textbf{\# DML training pairs} & \textbf{RC} & \textbf{MSE} \\ \hline
\textbf{C+L (F) \cite{parmar2017learning}}                        & NA                             & 0.64                      & 96.89                      \\ \hline
\textbf{\begin{tabular}[c]{@{}c@{}}Without DML \\  With Expert Bias\end{tabular}}      & NA                           & 0.56                     & 116.37                 \\ \hline

\textbf{\begin{tabular}[c]{@{}c@{}}With DML + Expert Bias \\  (without augmentation)\end{tabular}}      & 3000                           & 0.57                      & 138.29                    \\ \hline
\textbf{\begin{tabular}[c]{@{}c@{}}With DML + Expert Bias \\  (few augmented videos)\end{tabular}} & 14000                          & 0.626                    & 107.8                     \\ \hline
\textbf{\begin{tabular}[c]{@{}c@{}}With DML\\  (large augmentation)\end{tabular}}         & 26000                          & \textbf{0.69}             & \textbf{88.59}             \\ \hline
\end{tabular}
\end{table}

\begin{table*}[ht!]
\centering
\caption{Results of fine tuning the diving model on gymnastic vaults; C: C3D, L: LSTM, F: Final Label Scoring, I: Incremental Label Scoring, S: SVR}
\label{vault}
\begin{tabular}{|c|c|c|c|c|c|c|c|c|c|c|}
\hline
\textbf{}              & \multicolumn{8}{c|}{\textbf{Parmar \etal \cite{parmar2017learning}}}                                                                                                                                                                                                   & \multicolumn{2}{c|}{\textbf{Ours}}                \\ \hline
\multicolumn{1}{|l|}{} & \multicolumn{2}{l|}{\textbf{C + L (F)}}                              & \multicolumn{2}{l|}{\textbf{C + L (I)}}                              & \multicolumn{2}{l|}{\textbf{C + L + S (F)}} & \multicolumn{2}{l|}{\textbf{C + L + S (I)}} & \multicolumn{2}{l|}{\textbf{Best Performance Bias}}    \\ \hline
\textbf{Splits}        & \multicolumn{1}{l|}{\textbf{RC}} & \multicolumn{1}{l|}{\textbf{MSE}} & \multicolumn{1}{l|}{\textbf{RC}} & \multicolumn{1}{l|}{\textbf{MSE}} & \textbf{RC}         & \textbf{MSE}         & \textbf{RC}          & \textbf{MSE}         & \textbf{RC}   & \multicolumn{1}{l|}{\textbf{MSE}} \\ \hline
1                      & 0.39                             & 0.58                              & 0.52                             & 0.88                              & 0.47                & 0.36                 & 0.45                 & 0.77                 & 0.37          & 0.36                              \\ \hline
2                      & 0.45                             & 0.62                              & 0.63                             & 0.86                              & 0.55                & 0.5                  & 0.67                 & 1.04                 & 0.39          & 0.46                              \\ \hline
3                      & 0.34                             & 0.67                              & 0.34                             & 0.84                              & 0.44                & 0.52                 & 0.4                  & 0.83                 & 0.49          & 0.57                             \\ \hline
4                      & 0.19                             & 0.88                              & 0.33                             & 0.72                              & 0.16                & 0.61                 & 0.35                 & 0.64                 & 0.36         & 0.45                              \\ \hline
5                      & 0.35                             & 0.68                              & 0.56                             & 0.97                              & 0.34                & 0.58                 & 0.54                 & 0.92                 & 0.44          & 0.34                              \\ \hline
\textbf{Average}       & 0.34                             & 0.69                              & 0.47                             & 0.85                              & 0.39                & 0.51                 & 0.48                 & 0.84                 & \textbf{0.41} & \textbf{0.44}                      \\ \hline
\end{tabular}
\end{table*}

  \begin{table*}[h!]
    \caption{Clip Level contribution to the final score}\label{fig:feedback}
  \centering
  \begin{tabular}{ | c | m{4cm}| m{6cm} |m{4cm}| }
    \hline
    Video Number & Commentary & Clipwise similarity & Graph Interpretation\\ \hline
    214
    &
    $<$sos$>$ eight verse three and a half somersault oh he's a little short on the entry they're beautiful through the year as always is that top Chinese divers see he just didn't kick out early enough and {\color{red}\textbf{he's not vertical}} at all although {\color{red}\textbf{somehow he manages to make such a small splash}} that's great control underneath the water to pull that bubble down with him like many of the divers here they did take the chance to get a feel of the $<$eos$>$ 
    
    & 
    \begin{minipage}{.3\textwidth}
      \includegraphics[scale=0.35]{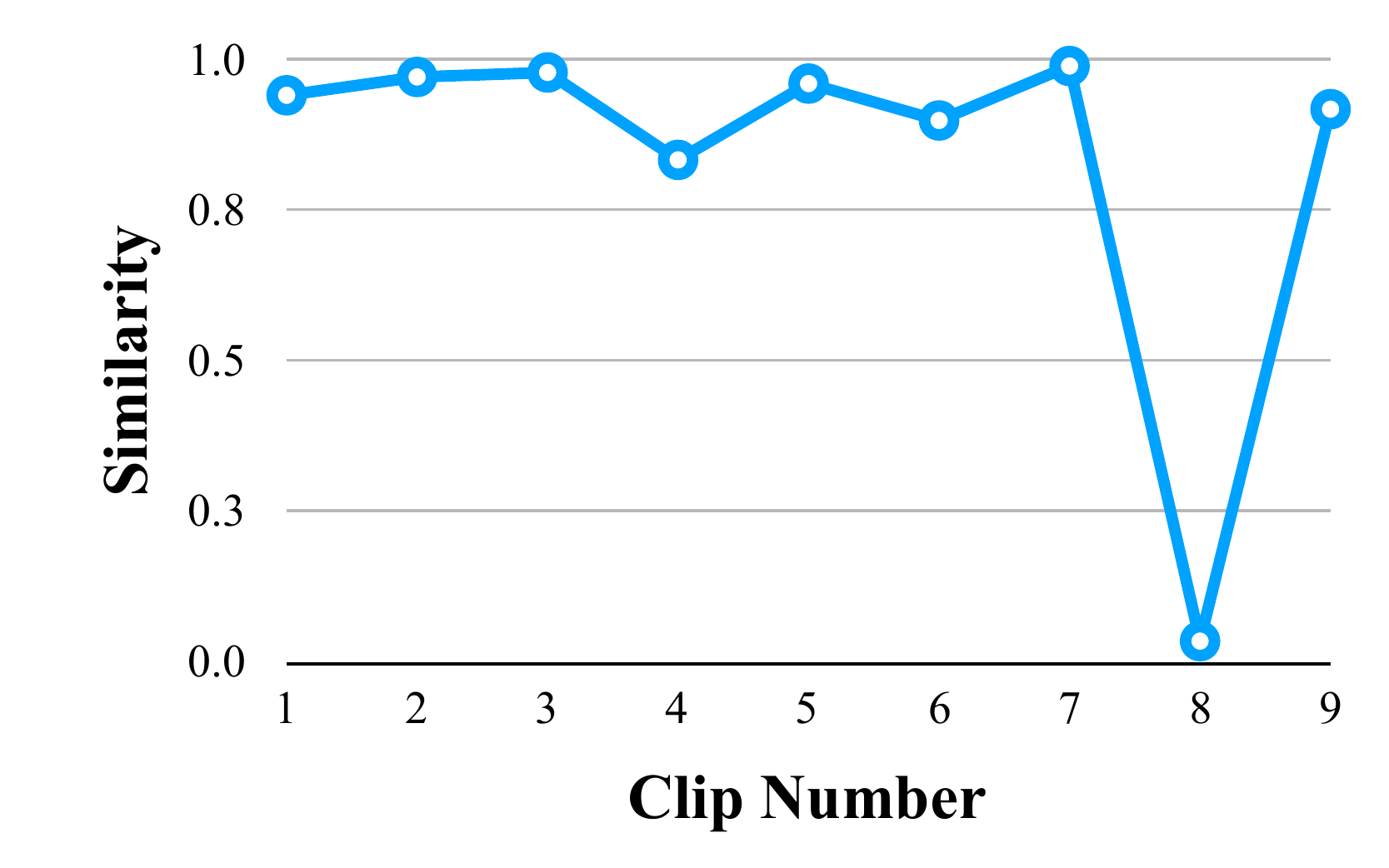}
    \end{minipage}
      
      & 
      \begin{itemize}
        \item Clip 8 - Entry Clip
        \item Clip 9 - Splash
        \item Similarity decreases on entry and increases in the next clip 
      \end{itemize}
    \\ \hline
    
    118
    &
    $<$sos$>$ {\color{red}\textbf{beautiful is back to near his best}} [Applause] [Music] three point two that's like it he loves to smile for the camera doesn't he that he loves this stage well he is a showman and to be fair even during the difficulties last night he came up smiling and waving at the camera I think he's one of those athletes Roger he just finds it a joy to be here wow I'm at the Olympics he's old he's done it all before but he just thrives on it and in 86 4 for Matthew he picked up another  $<$eos$>$ 
    
    & 
    \begin{minipage}{.3\textwidth}
      \includegraphics[scale=0.35]{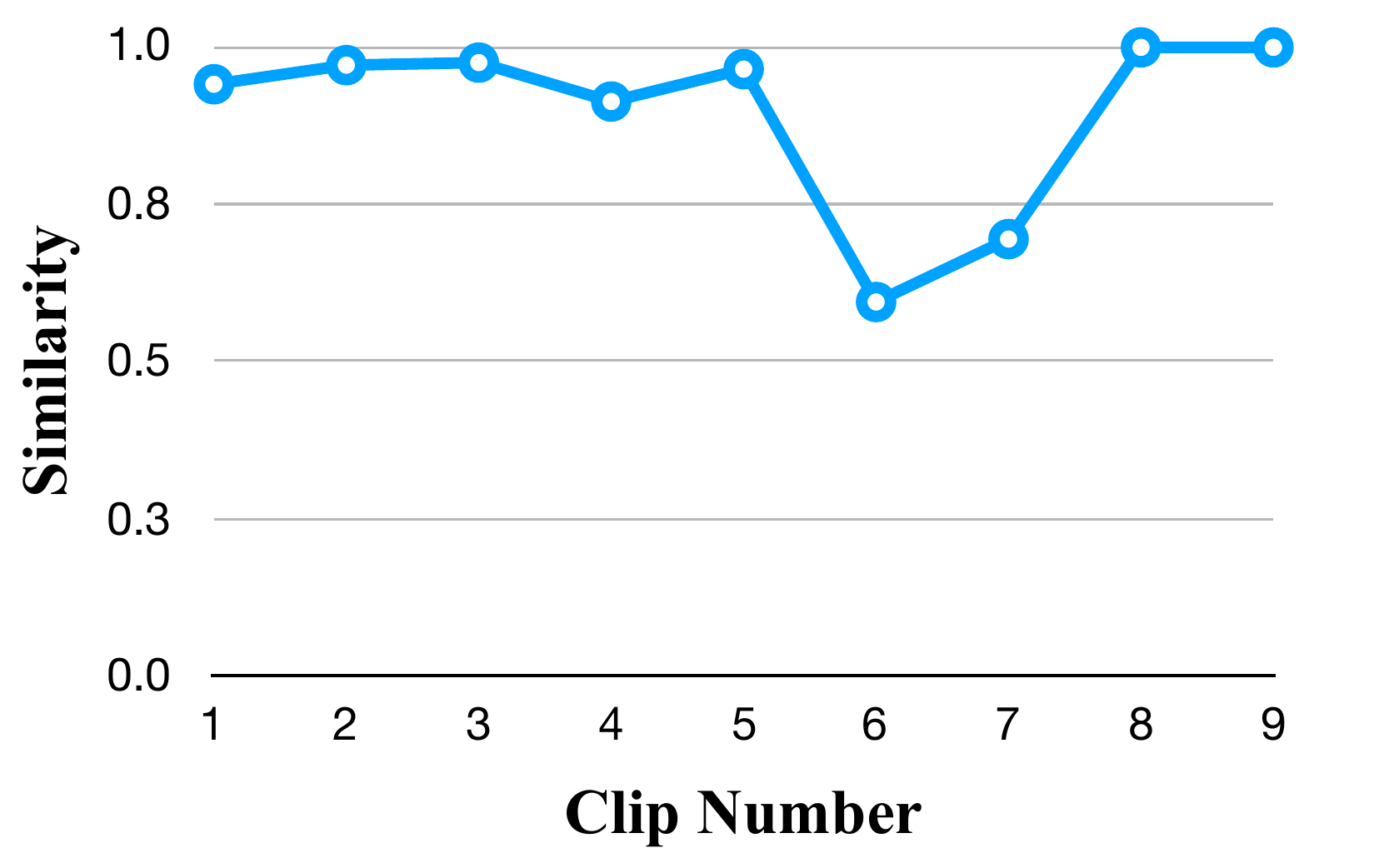}
    \end{minipage}

      & The similarities of the clip to the expert clip is more than 0.5 in all cases which indicates that the performance is overall good.
      
      \\ \hline
      
      44
    &
    $<$sos$>$ {\color{red}\textbf{it's a bit of a crash-landing}} I'm sad to say because Wolfram was having a good competition especially with his opening couple of dives and it's just gone missing for him confidence not there now short on position coming into the water and a well that's counting him out a sixty-four point eight is a disaster in  $<$eos$>$ 
    
    & 
    \begin{minipage}{.3\textwidth}
      \includegraphics[scale=0.35]{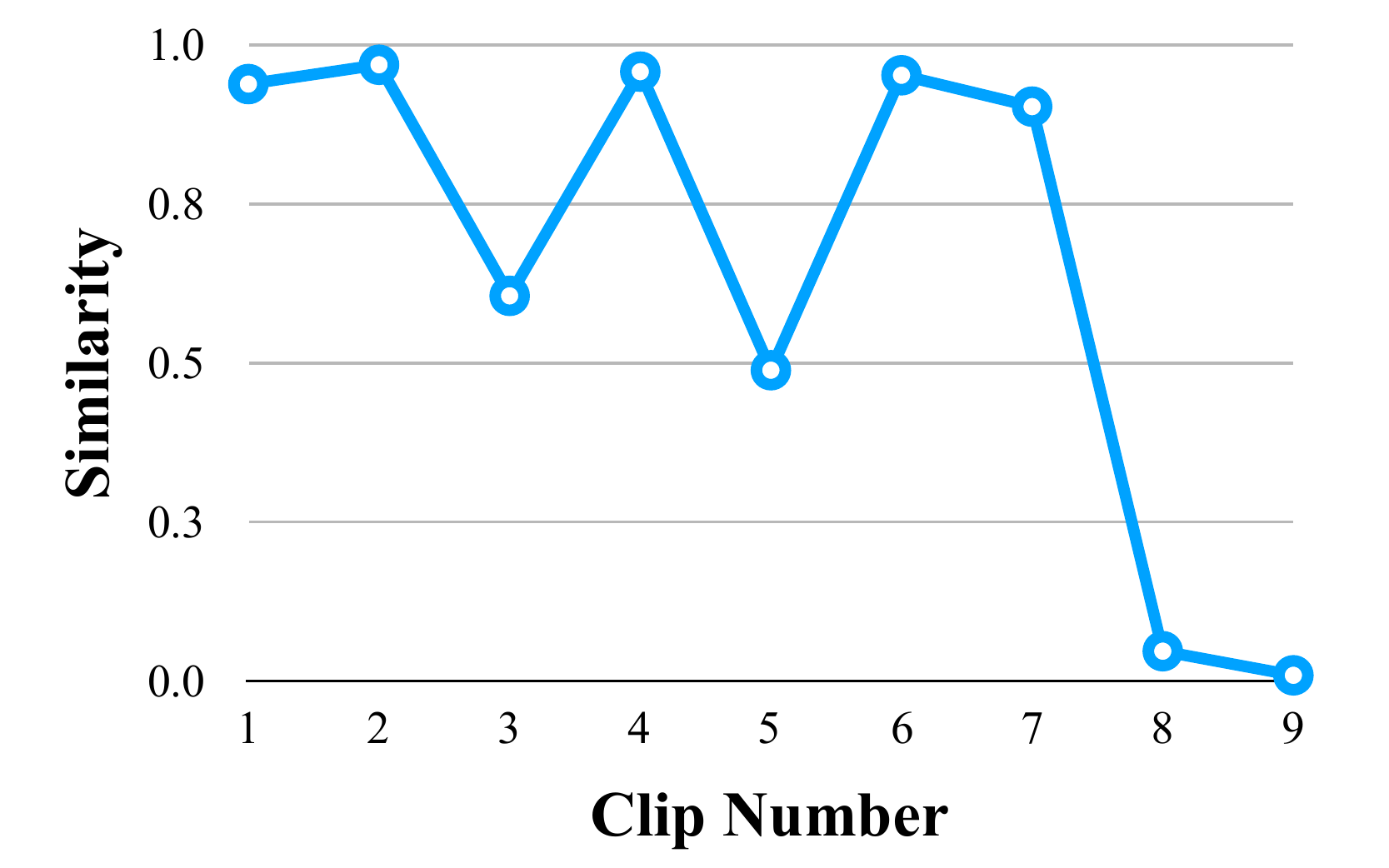}
    \end{minipage}

      & \begin{itemize}
        \item Clip 8 - Entry Clip
        \item Clip 9 - Splash
        \item The similarity decreases at the entry and splash clip 
      \end{itemize}
      
      \\ \hline

      292
    &
    $<$sos$>$ and the crowd gasps and we'll just leave that one to the judges because you saw as well as we did he would have seen that a diver before him duomo in a similar position in terms of rankings make a mistake and he had a chance to capitalize just made him move up that one more spot but if anything {\color{red}\textbf{he's put in a worst dive}} 39 6s Gilson so tough  $<$eos$>$ 
    
    & 
    \begin{minipage}{.3\textwidth}
      \includegraphics[scale=0.35]{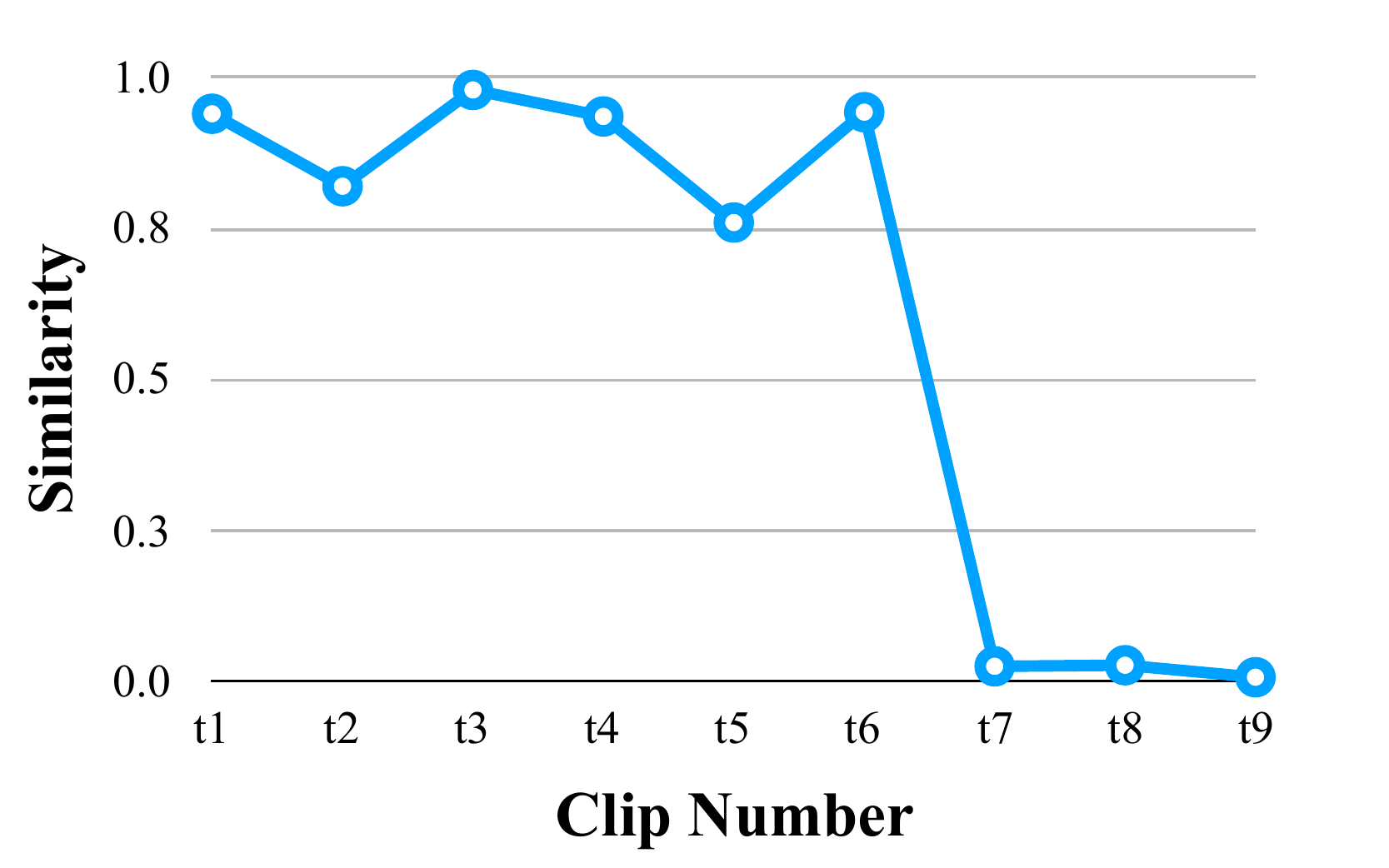}
    \end{minipage}

      & 
      Initial clips are padded ones and the similarity decreases in the last three clips indicating an overall performance.
      \\ \hline
  \end{tabular}
  
\end{table*}

\begin{table}[]
\centering
\caption{Perfomance comparison of feedback proposal schemes}
\label{tab:pr_feedback}
\begin{tabular}{|c|c|c|}
\hline
                          & \textbf{Precision} & \textbf{Recall} \\ \hline
\textbf{C3D + LSTM~\cite{parmar2017learning}}       & 0.18               & 0.52            \\ \hline
\textbf{C3D + LSTM + SVR~\cite{parmar2017learning}} & 0.11               & 0.36            \\ \hline
\textbf{Ours}             & 0.58               & 0.95            \\ \hline
\end{tabular}
\end{table}

    \subsection{Score Estimation Results}
    \textbf{\textit{Diving Results}}
    
    Tables~\ref{tab1} and \ref{tab2} give the average and split-wise Rank correlation and Mean square error results for different baseline models. 
    
    For comparing the scoring model with the other baselines, we evaluate three variants of our trained models where the biases (reference videos) were fixed as follows: 1) an expert bias of the first dive type is included in the pairs for all dive types; 2) dive specific expert performer bias is included in the pairs corresponding to the same dive type 3) dive specific worst performer bias is included in the pairs corresponding to the same dive type. Tables~\ref{tab1} and \ref{tab2} illustrate that the bias from the dive specific expert videos outperforms the other two variants. 
    

Table~\ref{e_w_dive} gives the number of the best and the worst performers for the Diving dataset and their respective scores for individual dive types. It is seen that the experts are more as compared to the worst performers. Thus we have around $380$ pairs to train the scoring model in case of expert bias and $174$ pairs in case of worst performer bias. This  leads to an improved efficiency of scoring in case of expert bias as compared to the worst bias even when both lead to coherent Siamese embedding. 
Further, including a constant bias for the task of scoring gives a poor performance as compared to the case when we use a worst or expert performer as show in Table~\ref{tab1} and \ref{tab2} where we choose expert of dive type $1$ as the constant. This trend is followed even when we choose all $4$ experts of dive type $3$ which gives us around $700$ input pairs. We achieve a still lower rank correlation of $0.62$ and higher mean square error of $98.64$ on average for the $10$ split.

    Further our proposed approach that constructs a video representation using the dive specific expert reference outperforms the traditional scoring models. 
    Our model has the maximum rank correlation and the minimum mean square error. 
    The improvement in the performance of our assessment model is because we make use of a representation that is specific to the dive type. 
    It is the new embedding that considers an expert bias with the candidate video. We also learned different regression models (C3D + LSTM) for different dive types and it was seen that 
    when compared to the use of bias-specific embeddings (our model), 
    the rank correlation between the predicted scores from the 6 different regression models and the ground-truth scores is lower 
    and the mean square error is higher. This result is indicated in Table~\ref{tab2}. 
    
    We also compare the performance of our model and the baseline models on evaluation of scores for the individual dive types. Table~\ref{tabb} illustrates that our model outperforms the baseline models for $5$ dive types and performs comparable for the $6th$ dive type too. Thus even when the DML module was not trained for $3$ of the $6$ dive types and the model has shown good generalization. 
    
    Next we check how the distance similarity metric learning impacts the performance of scoring. We skip the first phase of our framework \textit{i.e.} the DML module and directly perform scoring relative to expert. Table~\ref{fig:tab3} shows that without the DML phase the model gives a bad scoring performance. Following this the DML module is introduced and trained with and without data augmentation. 
    %
    With no augmentation there are too few ($1300$) pairs and the  Metric Learning performance is poorer compared to when we include more pairs using augmentation($26000$).  Further, a bad Metric Learning adversely impacts the task of scoring. It is seen that with too few DML module training pairs, the performance of the scoring is lower than the original C3D-LSTM \cite{parmar2017learning}. However, with data augmentation there is a positive impact on the performance of scoring and the DML module can identify intra-class variations. 
    
    Finally we compare the performance of our model with the C3D-LSTM model \cite{parmar2017learning} for gymnastic vaults. The other two baselines are not developed for gymnastic vaults. 
    
    \textbf{\textit{Vault Results}}
    
    We fine-tuned our model and the C3D+LSTM model developed by Parmar \etal~\cite{parmar2017learning} for diving to that of gymnastic vaults. Table~\ref{vault} shows that our model performs better in terms of both Rank Correlation and Mean Square Error than most of the C3D+LSTM model variations introduced by Parmar \etal \cite{parmar2017learning}. This shows that the distance metric learning is generic in terms of actions being compared. Thus our model not only performs better for the dive types not included in the DML training, but also gives better performance for gymnastic vaults after fine tuning. 
  \begin{figure*}%
\centering
\subfigure[][]{%
\label{fig:ex3-a}%
\includegraphics[scale=0.2]{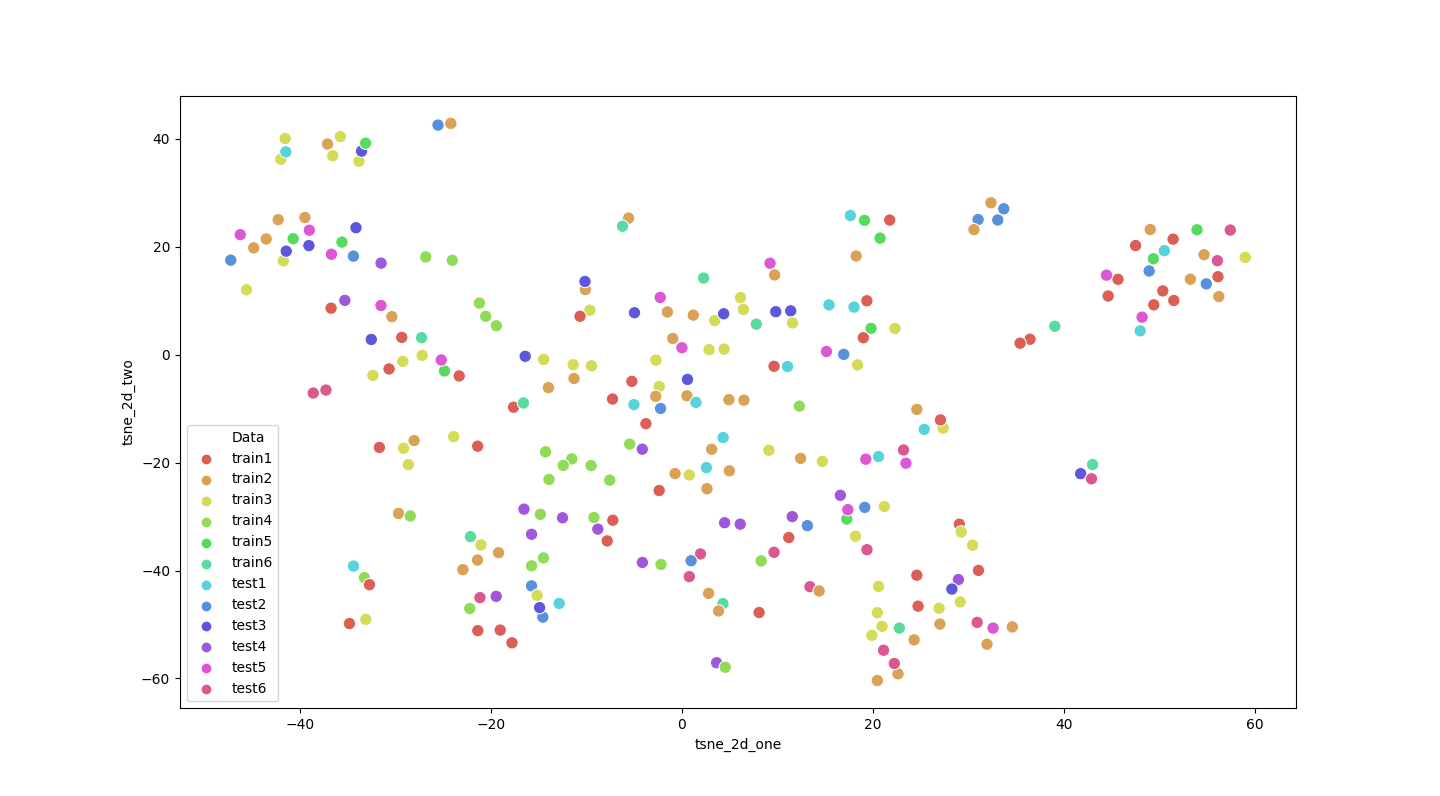}}%
\hspace{4pt}%
\subfigure[][]{%
\label{fig:ex3-b}%
\includegraphics[scale=0.2]{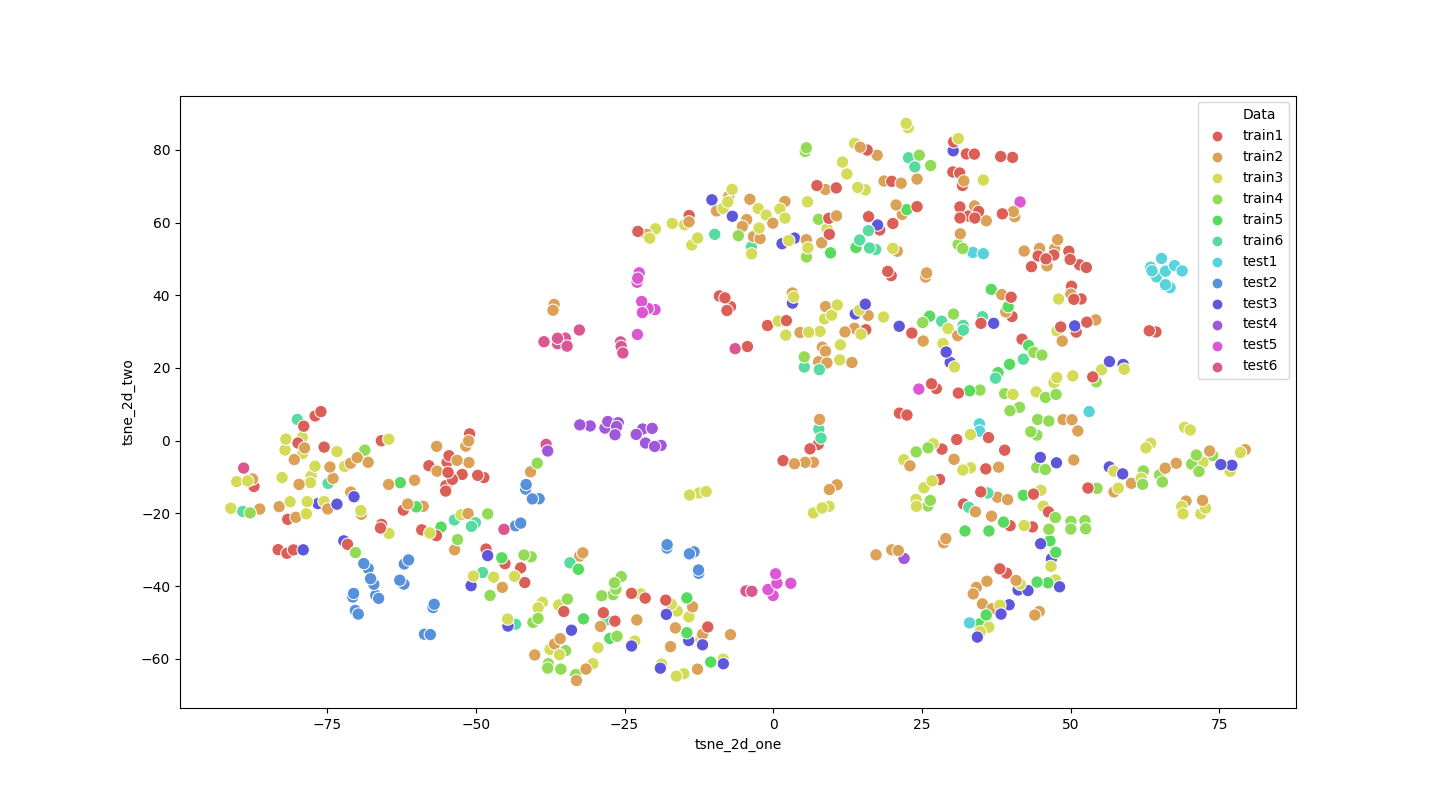}} \\
\subfigure[][]{%
\label{fig:ex3-c}%
\includegraphics[scale=0.2]{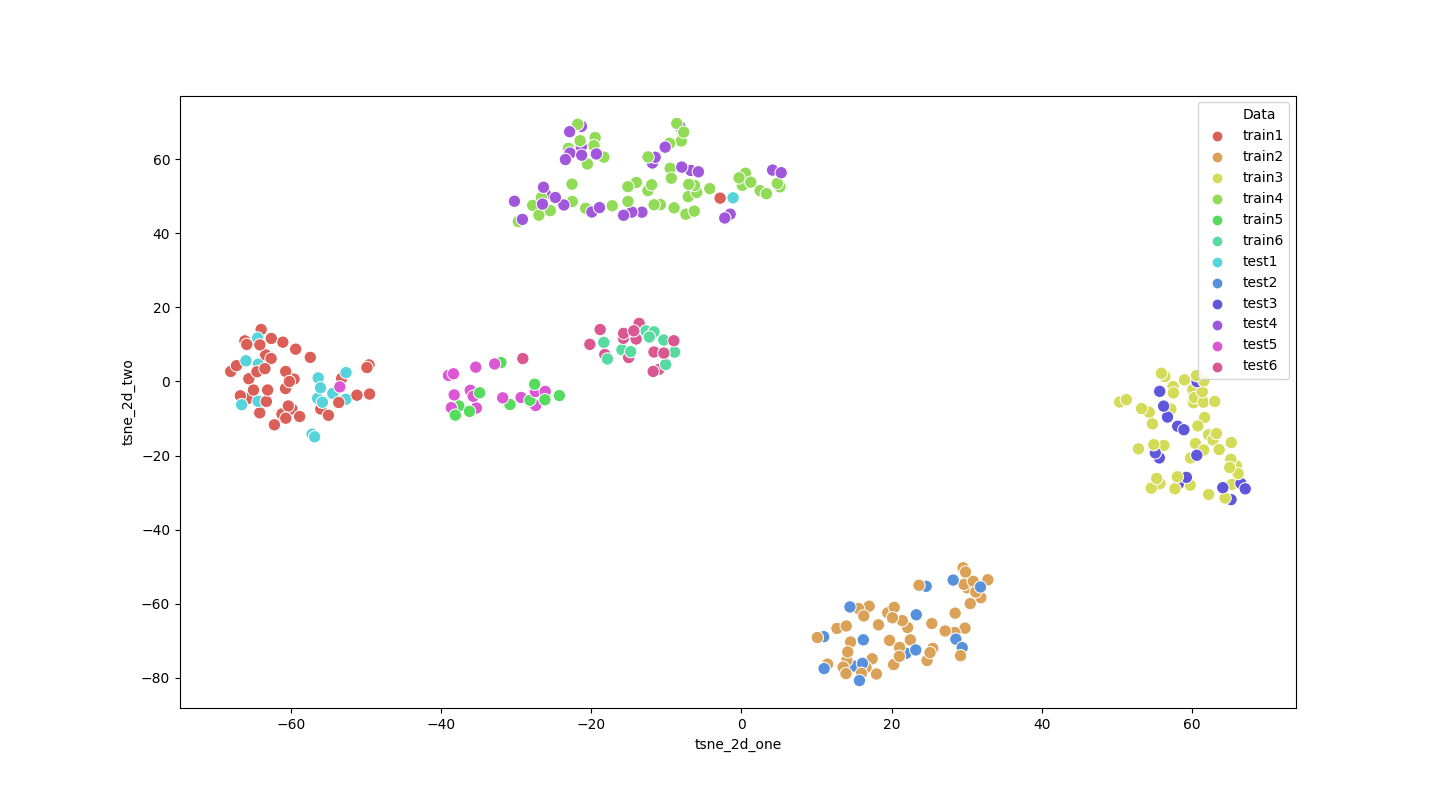}}%
\hspace{4pt}%
\subfigure[][]{%
\label{fig:ex3-d}%
\includegraphics[scale=0.2]{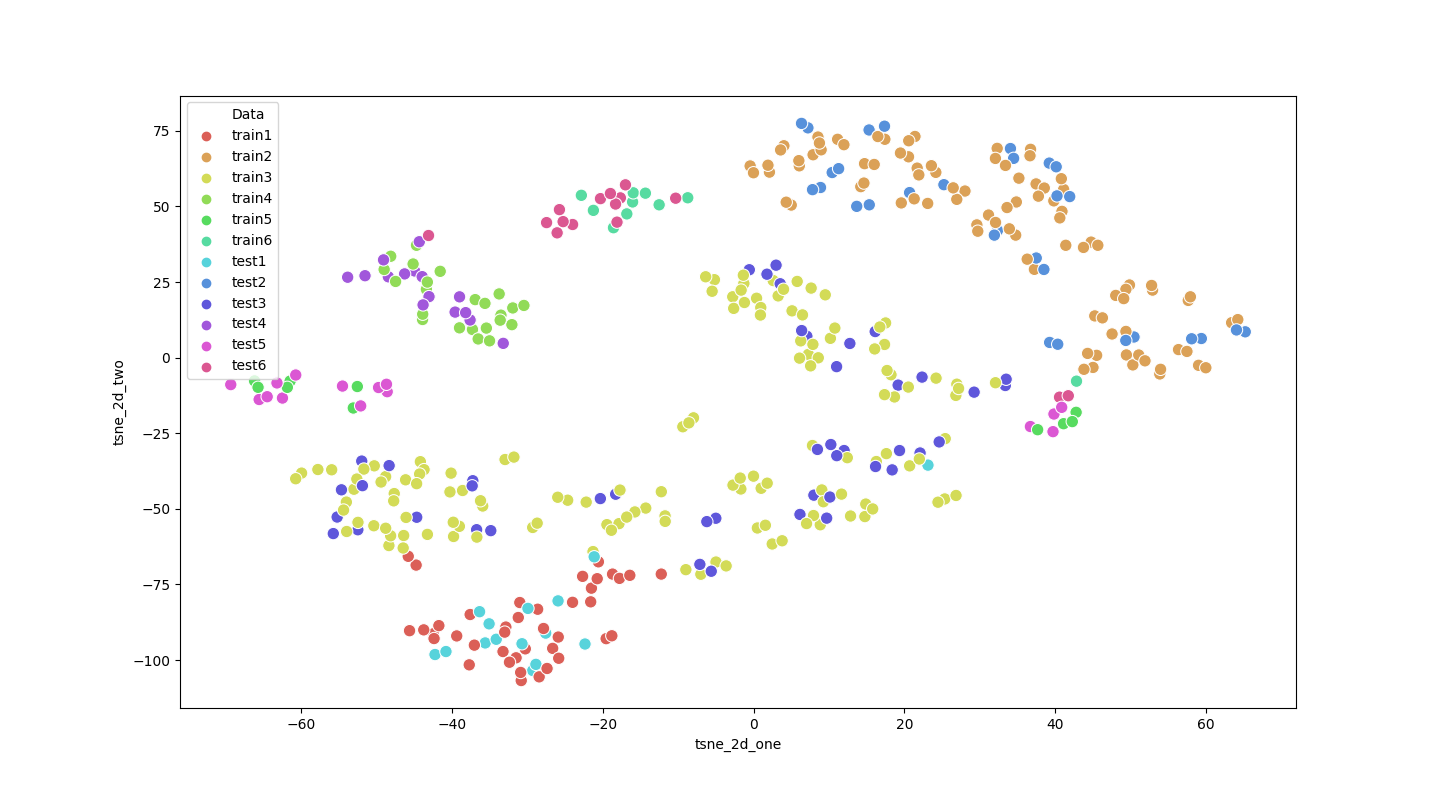}}%
\caption[t-SNE Plots of LSTM embedding and Siamese embedding of train and test data]{t-SNE Plots of LSTM embedding and Siamese embedding of train and test data:
\subref{fig:ex3-a} the LSTM embedding $O_p$;
\subref{fig:ex3-b} Siamese Embedding $D_2$ with Constant Bias at $O_p$;
\subref{fig:ex3-c} Siamese Embedding $D_2$ with Worst Performer Bias at $O_p$; and,
\subref{fig:ex3-d} Siamese Embedding $D_2$ with Expert Performer Bias at $O_p$}%
\label{fig:ex3}%
\end{figure*}
  
    \subsection{Feedback Proposals}
Here we illustrate the efficacy of the Distance Metric Learning module in predicting the contributions of each clip of the action to the final score. Following the strategy of clip wise similarity computation proposed in the last section, we achieve a reasonably good feedback for the actions that correlate with the Olympics commentaries, which are provided by Parmar \etal \cite{parmar2019and}. Table\ref{fig:feedback} shows the feedback proposals for some videos. We see that the similarity of the clips decreases in consistency with the video's commentary.

We propose a scheme to compare the performance of our clip-level feedback to that provided by Parmar \etal~\cite{parmar2017learning}. For the test videos, the commentaries provided in the Multitask AQA diving dataset \cite{parmar2019and} are interpreted to find the faulty clips in the performance. This is considered as the ground truth for feedback proposals. 
The clip wise similarity computed from our DML module signify faulty clips when the Sigmoid output is less than $0.5$ \textit{i.e.} the clips are not similar to the corresponding clip of an expert execution.

We evaluate the precision and recall for the two systems in identifying the faulty clips. Table~\ref{tab:pr_feedback} lists the comparison results for the two systems. Our system outperforms the LSTM+C3D aggregations proposed by Parmar \etal~\cite{parmar2017learning}. The high recall of our scheme signifies fewer faulty clips that are missed. On the contrary the lower precision mostly occur due to misalignment of the clips.

\subsection{Analysis of various embedding} 

We analyse the LSTM embedding $O_p$ and the Siamese embedding $D_2$ (Figure~\ref{fig:siamese}, with weights learned during the DML module training. The 2D t-SNE plots of these embeddings for training and test data are plotted to evaluate their coherence across various dive types. Figure~\ref{fig:ex3-a} shows the plots of the LSTM embedding. It is seen that the different dive types do not form clear clusters with the LSTM embedding. Similar non-distinct clusters are observed with the Siamese embedding $D_2$ when a constant expert of dive type $3$ is passed as an input in all the pairs (Figure~\ref{fig:ex3-b}).

However, when the bias $O_p$ is made dive-specific by either including the worst performer per dive or the best performer per dive, the Siamese embedding are seen to become coherent for individual dive types (Figure~\ref{fig:ex3-c} and \ref{fig:ex3-d}). Thus an inclusion of a bias $O_p$ specific to the dive types and using the $D_2$ embedding for the task of scoring is a good choice, which is exhibited by improved performance as discussed in Table~\ref{tab1} and \ref{tab2}.

\section{Conclusions}
\label{sec:conclusion}
Human action quality assessment or action scoring has been posed as a regression problem in the past works. The simple regression-based solutions lack interpretability. We introduce a novel approach  for action scoring in which the performances are compared with the reference videos to estimate the final score. A deep metric learning module learns the similarity metric between two videos using the difference in the scores of the videos. This module is then used to find the similarity of the video to the reference video. Such an approach can capture and assess the the temporal variations of videos with the reference video. Our experiments on Olympics Diving and Gymnastic vaults actions show that the proposed approach that includes a bias of the high-rated reference video performs better than the traditional scoring methods. Further, we introduce an unsupervised technique to provide the sub-action level feedback in order to make the scores more explicable. Use of deep learning metric to assess long term actions where alignment can be a major issue is the future direction of our work.



\ifCLASSOPTIONcaptionsoff
  \newpage
\fi

\end{document}